\documentclass{article} % For LaTeX2e
\usepackage{iclr2021_conference}
\usepackage{times}

\usepackage[utf8]{inputenc} % allow utf-8 input
\usepackage[T1]{fontenc}    % use 8-bit T1 fonts
\usepackage{hyperref}       % hyperlinks
\usepackage{url}            % simple URL typesetting
\usepackage{booktabs}       % professional-quality tables
\usepackage{amsfonts}       % blackboard math symbols
\usepackage{nicefrac}       % compact symbols for 1/2, etc.

\usepackage{microtype}      % microtypography
\usepackage{multirow}
\usepackage[linesnumbered,ruled]{algorithm2e}
\usepackage[title]{appendix}
\usepackage{amssymb}
\usepackage{csquotes}
\usepackage{bbm}
\usepackage{amsthm}
\usepackage{graphicx} % DO NOT CHANGE THIS
\usepackage{algorithmic}
\usepackage{tablefootnote}
\usepackage{bm}

\usepackage{thm-restate}

\usepackage{array}
\newcolumntype{P}[1]{>{\centering\arraybackslash}p{#1}}

\newcommand{\norm}[1]{\left\lVert#1\right\rVert}
\newcommand{\idot}[1]{\langle#1\rangle}

\newtheorem{theorem}{Theorem}[section]

\newtheorem{lemma}[theorem]{Lemma}
\newtheorem{remark}{Remark}
\newtheorem{definition}{Definition}

\usepackage{amsmath,amsfonts,bm}

\def\eqref#1{equation~\ref{#1}}
\def\1{\bm{1}}

\def\vxi{{\bm{\xi}}}

\def\vg{{\bm{g}}}

\def\vr{{\bm{r}}}

\def\vu{{\bm{u}}}
\def\vv{{\bm{v}}}
\def\vw{{\bm{w}}}
\def\vx{{\bm{x}}}

\def\vz{{\bm{z}}}

\def\mA{{\bm{A}}}
\def\mB{{\bm{B}}}

\def\mG{{\bm{G}}}

\def\mI{{\bm{I}}}

\def\mM{{\bm{M}}}

\def\mR{{\bm{R}}}
\def\mS{{\bm{S}}}

\def\mV{{\bm{V}}}
\def\mW{{\bm{W}}}
\def\mX{{\bm{X}}}

\def\mSigma{{\bm{\Sigma}}}

\DeclareMathAlphabet{\mathsfit}{\encodingdefault}{\sfdefault}{m}{sl}
\SetMathAlphabet{\mathsfit}{bold}{\encodingdefault}{\sfdefault}{bx}{n}

\def\gP{{\mathcal{P}}}

\def\sC{{\mathbb{C}}}
\def\sD{{\mathbb{D}}}
\def\sF{{\mathbb{F}}}

\def\sM{{\mathbb{M}}}

\def\sP{{\mathbb{P}}}

\def\sR{{\mathbb{R}}}

\def\tvv{{\tilde{\vv}}}
\def\tvu{{\tilde{\vu}}}

\newcommand{\E}{\mathbb{E}}

\usepackage{hyperref}
\usepackage{url}

\iclrfinalcopy

\title{Do not Let Privacy Overbill Utility:  Gradient Embedding Perturbation for Private Learning}

\author{Da Yu$^{1,2,*}$, Huishuai Zhang$^{2,}$\thanks{Authors contribute equally to this work.}\,\,, Wei Chen$^{2}$, Tie-Yan Liu$^{2}$\\
$^1$School of Computer Science and Engineering, Sun Yat-sen University \\
$^2$Microsoft Research Asia\\
$^1$\texttt{yuda3@mail2.sysu.edu.cn}\\ 
$^2$\texttt{\{huzhang,wche,tyliu\}@microsoft.com}
}

\begin{document}

\maketitle

\begin{abstract}

The privacy leakage of the model about the training data can be  bounded in the differential privacy mechanism. However, for meaningful privacy parameters, a differentially private model degrades the utility drastically when the model comprises a large number of trainable parameters.  In this paper, we propose an algorithm  \emph{Gradient Embedding Perturbation (GEP)} towards training differentially private deep models with decent accuracy. Specifically, in each gradient descent step, GEP first projects individual private gradient into a non-sensitive anchor subspace, producing a low-dimensional gradient embedding and a small-norm residual gradient. Then, GEP perturbs the low-dimensional embedding and the residual gradient separately according to the privacy budget. Such a decomposition permits a small perturbation variance, which greatly helps to break the dimensional barrier of private learning. With GEP, we achieve decent accuracy with reasonable computational cost and modest privacy guarantee for deep models.  Especially, with privacy bound $\epsilon=8$, we achieve $74.9\%$ test accuracy on CIFAR10 and $95.1\%$ test accuracy on  SVHN, significantly improving over existing results. 

\end{abstract}

\section{Introduction}

Recent works have shown that the trained model  may leak/memorize the information of its training set \citep{fredrikson2015model, wu2016methodology, shokri2017membership, hitaj2017deep}, which  raises privacy issue  when the models are trained with sensitive data. \emph{Differential privacy} (DP) mechanism  provides a way to quantitatively measure and upper bound such information leakage. It theoretically ensures that the influence of any individual sample is negligible  with the DP parameter $\epsilon$ or $(\epsilon, \delta)$. Moreover, it has been observed that  differentially private models can also resist model inversion  attack \citep{carlini2019secret}, membership inference attack \citep{rahman2018membership,bernau2019assessing,sablayrolles2019white,yu2021how}, gradient matching attack \citep{zhu2019deep}, and data poisoning attack \citep{ma2019data}.

\begin{figure}

    \begin{minipage}{0.4\textwidth}
        \centering
        \includegraphics[width=0.7\linewidth]{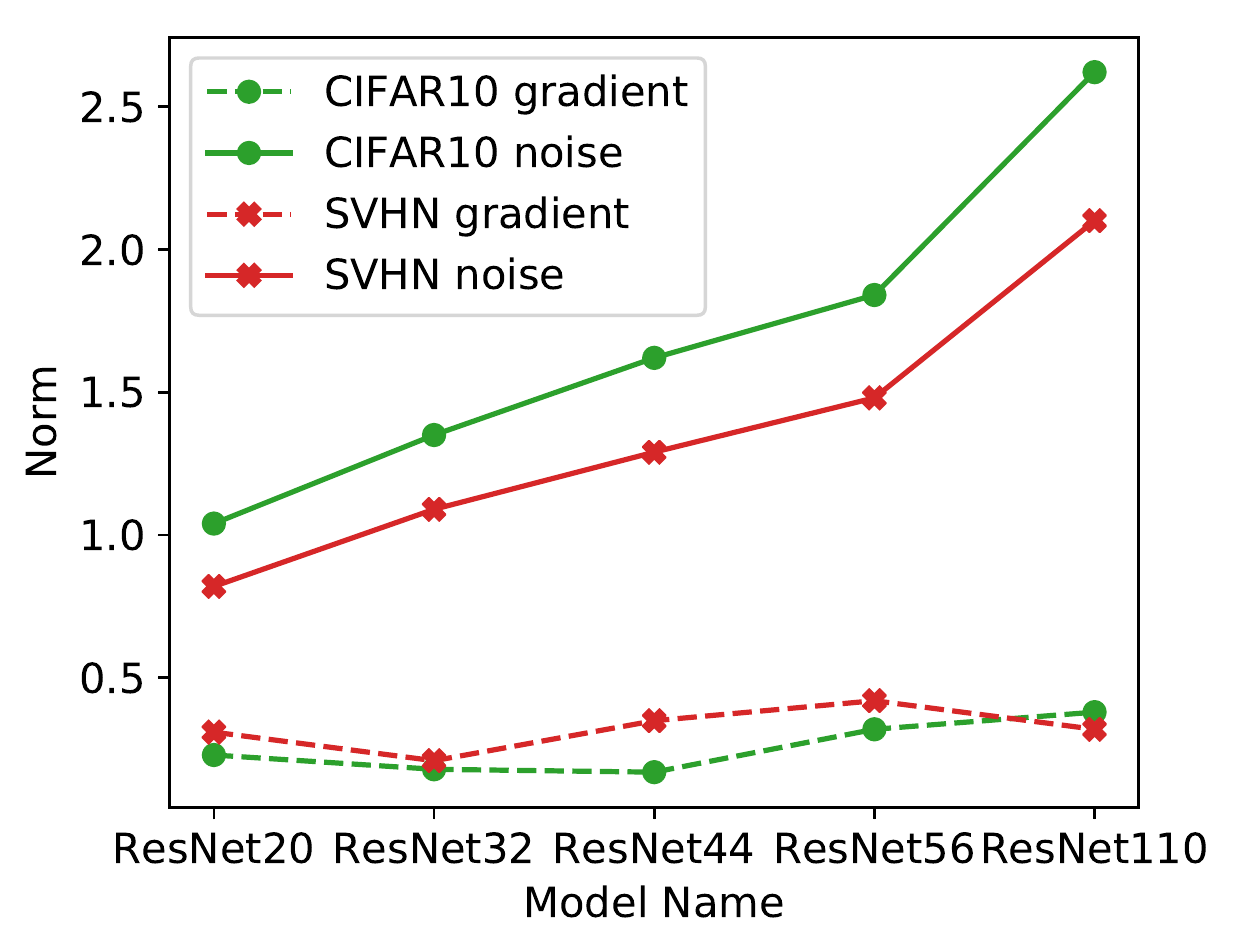}
          \caption{Noise  norm vs gradient  norm of ResNet20 at initialization.  The noise variance is chosen such that SGD satisfies $(5,10^{-5})$-DP after $90$ epochs in \cite{abadi2016deep}.  }
          \label{fig:grad_noise}
        
    \end{minipage}
    \hfill
    \begin{minipage}{0.4\textwidth}
        \centering
        \includegraphics[width=0.65\linewidth]{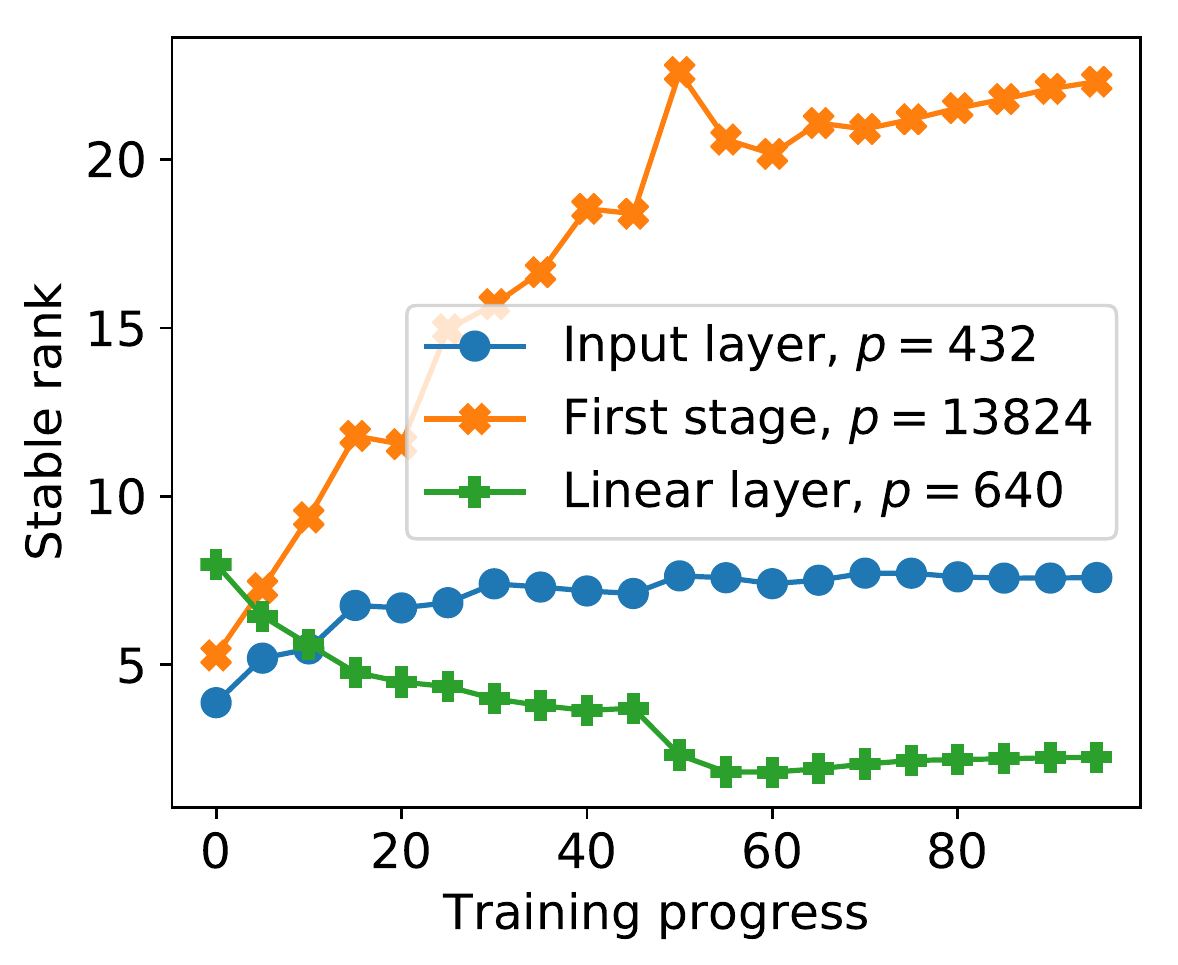}
        \caption{Stable rank $\|\cdot\|_F^2/\|\cdot\|^2$  \citep{tropp2015introduction} of batch gradient matrix of given groups (with $p$ parameters). The setting is ResNet20 on CIFAR-10.   The stable rank is small throughout training. } %(the smaller the value, the closer to the input layer)
        \label{fig:stbl_rank}
        
    \end{minipage}    
\end{figure}

One popular way to achieve differentially private machine learning is to perturb the training process with noise  \citep{song2013stochastic,bassily2014differentially,shokri2015privacy,wu2017bolt,fukuchi2017differentially,iyengar2019towards,phanscalable}. Specifically, \emph{gradient perturbation} perturbs the gradient at each iteration of (stochastic) gradient descent algorithm and guarantees the privacy of the final model via \emph{composition property} of DP. It is worthy to note that gradient perturbation does not assume (strongly) convex objective and hence is applicable to various settings \citep{abadi2016deep,wang2017differentially,lee2018concentrated,jayaraman2018distributed,wang2019differentially,yu2020gradient}. Specifically, for given gradient sensitivity $S$, a general form of gradient perturbation is to add an isotropic Gaussian noise $\vz$  to the gradient $\vg \in\mathbb{R}^{p}$ independently for each step,
\begin{equation}
\begin{aligned}
\label{eq:general_gp}
\tilde \vg=\vg+\vz, \;\; \text{where} \;\; \vz\sim \mathcal{N}(0,\sigma^{2}S^{2}\mI_{p\times p}).
\end{aligned}
\end{equation}
One can set proper variance $\sigma^2$ to make each update differentially private with parameter $(\epsilon,\delta)$. It is easy to see that the intensity of the added noise $\E[\|\vz\|^2]$ scales linearly with the model dimension $p$. This indicates that as the model becomes larger, the useful signal, i.e., gradient, would be submerged in the added noise (see Figure~\ref{fig:grad_noise}). This dimensional barrier restricts the utility of deep learning models trained with gradient perturbation.

The dimensional barrier is attributed to the fact that the added noise is isotropic while the gradients live on a very low dimensional manifold, which has been observed in \citep{gur2018gradient, vogels2019powersgd,gooneratne2020low,li2020hessian} and is also verified in Figure~\ref{fig:stbl_rank} for the gradients of a 20-layer ResNet \citep{he2016deep}.  Hence to limit the noise energy, it is natural to think

``Can we reduce the dimension of gradients first and then add the isotropic noise onto a low-dimensional gradient embedding?"

\begin{figure}
    \centering
  \includegraphics[width=0.9\linewidth]{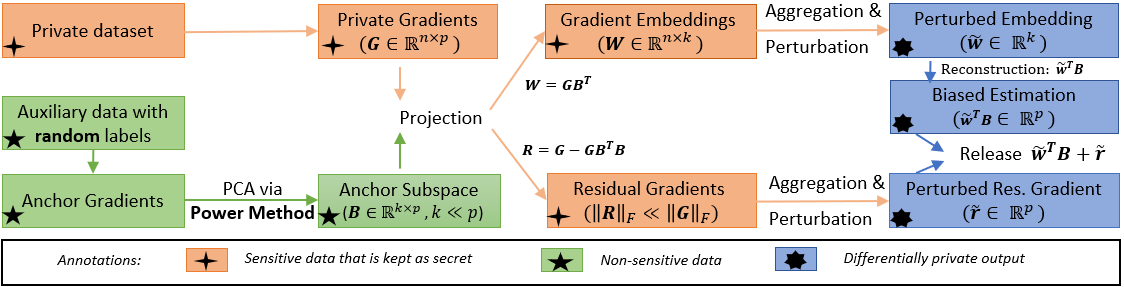}
  \caption{Overview of the proposed GEP approach. 1) We estimate an anchor subspace on some non-sensitive data; 2) We project the private gradients into the anchor subspace, producing low-dimensional embeddings and residual gradients; 3) We  perturb the gradient embedding and residual gradient separately to guarantee differential privacy.   The auxiliary data are only required to share similar features as the private data. In our experiments, we use $2000$ images from ImageNet as auxiliary data for MNIST, SVHN, and CIFAR-10 datasets. }
  \label{fig:overview}
\end{figure}

The answer is affirmative. We propose a new algorithm \emph{Gradient Embedding Perturbation (GEP)}, illustrated in Figure~\ref{fig:overview}. Specifically, we first compute \emph{anchor gradients} on some non-sensitive auxiliary data, and identify  an \emph{anchor subspace} that is spanned by several top principal components of the anchor gradient matrix. Then we project the private gradients into the anchor subspace and obtain  low-dimensional \emph{gradient embeddings} and small-norm \emph{residual gradients}. Finally, we perturb the gradient embedding and residual gradient separately according to the sensitivities and privacy budget. 

We intuitively argue why GEP could reduce the perturbation variance and achieve good utility for large models. First, because the gradient embedding has a very low dimension, the added isotropic noise on embedding has small energy that scales linearly only with the subspace dimension. Second, if the anchor subspace can cover most of the gradient information, the residual gradient, though high dimensional, should have small magnitude, which permits smaller added noise to guarantee the same level privacy because of the reduced sensitivity. Overall, we can use a much lower perturbation compared with the original gradient perturbation to guarantee the same level of privacy.

We emphasize several properties of GEP. First, the non-sensitive auxiliary data assumption is weak. In fact, GEP only requires a small number of non-sensitive unlabeled data following a similar feature distribution as the private data, which often exist even for learning on sensitive data. In our experiments, we use a few unlabeled  samples from ImageNet to serve as auxiliary data for MNIST, SVHN, and CIFAR-10. This assumption is much weaker than the public data assumption in previous works \citep{papernot2016semi, papernot2018scalable,alon2019limits,wang2020differentially}, where the public data should follow exactly the same distribution as the private data. Second, GEP produces an unbiased estimator of the target gradient because of  releasing both the perturbed gradient embedding and the perturbed residual gradient, which turns out to be critical for good utility. Third, we use \emph{power method} to estimate the principal components of anchor gradients, achievable  with a few matrix multiplications. The fact that GEP is not sensitive to the choices of subspace dimension further allows a very efficient implementation.

Compared with existing works of differentially private machine learning, our contribution can be summarized as follows:  (1) we propose a novel algorithm GEP that achieves good utility for large models with modest differential privacy guarantee; (2) we show that GEP returns an unbiased estimator of target private gradient with much lower perturbation variance than original gradient perturbation; (3) we demonstrate that GEP achieves state-of-the-art utility in differentially private learning with three benchmark datasets.  Specifically, for $\epsilon=8$, GEP achieves $74.9\%$ test accuracy on CIFAR-10 with a ResNet20 model. To the best of our knowledge, GEP is the first algorithm that can achieve such utility with training deep models from scratch for a ``single-digit" privacy budget\footnote{\cite{abadi2016deep}  achieve $73\%$ accuracy on CIFAR-10 but they need to pre-train the model on CIFAR-100.}.

%do not need to be sampled from the same distribution

\vspace{-1mm}
\subsection{Related work}

\label{sec:related_work}

\vspace{-2mm}

Existing works studying differentially private machine learning in high-dimensional setting can be  roughly categorized into two sets. One is treating the optimization of the machine learning objective as a whole mechanism and adding noise into this process. The other one is based on the  knowledge transfer of machine learning models, which trains a differentially private publishable student model with  private signals from teacher models.  We review them one by one.

Differentially private convex optimization in high-dimensional setting has been studied extensively over the years \citep{kifer2012private,thakurta2013differentially,talwar2015nearly,wang2019sparse,wang2019differentially}. Although  these methods demonstrate good utility on some convex settings, their analyses  can not be directly applied to non-convex setting. Right before the submission, we note two independent and concurrent works \citep{zhou2020bypassing,kairouz2020dimension} that also leverage the gradient redundancy to reduce the added noise. Specifically, \cite{kairouz2020dimension} track historical gradients to do dimension reduction for private AdaGrad. \cite{zhou2020bypassing} requires gradients on some public data and then project the noisy gradients into a public subspace at each update.  One core difference between  these two works and GEP is that we introduce residual gradient perturbation and GEP produces an unbiased estimator of the private gradients, which is essential for achieving the superior utility. Moreover, we weaken the auxiliary data assumption and introduce several designs that significantly boost the efficiency and  applicability of GEP. 

One recent progress towards training arbitrary models with differential privacy is \emph{Private Aggregation of Teacher Ensembles (PATE)} \citep{papernot2016semi, papernot2018scalable, jordon2019pate}. PATE first trains independent teacher models on disjoint shards of private data. Then it trains a student model with privacy guarantee by distilling noisy predictions of teacher models on some public samples. In comparison, GEP only requires some non-sensitive data that have similar natural features as the private data while PATE requires the public data follow  exactly the same distribution as the private data and in practice it uses a portion of the test data to serve as public data. Moreover, GEP demonstrates better performance than PATE especially for complex datasets,  e.g., CIFAR-10, because GEP can train the model with the whole private data rather than a small shard of data.

\vspace{-1mm}
\section{Preliminaries}

\label{sec:pre}
\vspace{-2mm}

We introduce  some notations and definitions. We use bold lowercase letters, e.g., $\vv$, and bold capital letters, e.g., $\mM$, to denote vectors and matrices, respectively. The $L^2$ norm of a vector $\vv$ is denoted by $\|\vv\|$. The spectral norm and the Frobenius norm of a matrix $\mM$ are denoted by $\|\mM\|$ and $\|\mM\|_F$, respectively. A sample $d=(\vx,y)$  consists of feature $\vx$ and label $y$. A dataset $\sD$ is a collection of individual samples. A dataset $\sD'$ is said to be a neighboring dataset of $\sD$ if they differ in a single sample, denoted as $\sD\sim \sD'$. Differential privacy ensures that the outputs of an algorithm on neighboring datasets have approximately indistinguishable distributions.  

\begin{definition}[$(\epsilon,\delta)$-DP \citep{dwork2006our, dwork2006calibrating}]
A randomized mechanism $\mathcal{M}$  guarantees $(\epsilon,\delta)$-differential privacy if for any two neighboring input datasets $\sD\sim \sD^{'}$ and for any subset of outputs $S$ it holds that $\text{Pr}[\mathcal{M}(\sD)\in S]\leq e^{\epsilon}\text{Pr}[\mathcal{M}(\sD^{'})\in S]+\delta$.
\end{definition}

By its definition, $(\epsilon, \delta)$-DP controls the maximum influence that any individual sample can produce.  One can adjust the privacy parameters to trade off between privacy and utility.  Differential privacy is immune to \emph{post-processing} \citep{algofound}, i.e., any function applied on the output of a differentially private algorithm would not increase the privacy loss as long as it does not have new interaction with the private dataset. Differential privacy also allows \emph{composition}, i.e., the composition of a series of differentially private mechanisms is also differentially private but with different parameters. Several variants of $(\epsilon, \delta)$-DP have been proposed  \citep{bun2016concentrated,dong2019gaussian} to address certain weakness of $(\epsilon, \delta)$-DP, e.g., they achieve better composition property. In this work, we use \emph{Rényi differential privacy} \citep{mironov2017renyi} to track the privacy loss and then convert it to $(\epsilon, \delta)$-DP.

Suppose that there is  a private dataset $\sD = \{(\vx_i, y_i)\}_{i=1}^n$ with $n$ samples. We want  to train a model $f$ to learn the mapping in $\sD$. Specifically, $f$ takes $\vx$ as input and outputs a label $y$, and $f$ has parameter $\theta\in \sR^p$. The training objective is to minimize an empirical risk $\frac{1}{n}\sum_{i=1}^n \ell(f(\vx_i), y_i)$, where $\ell(\cdot, \cdot)$ is a loss function. We further assume that there is an auxiliary dataset $\sD^{(a)}= \{(\tilde{\vx}_j, \tilde{y}_j)\}_{j=1}^m$ that $\tilde{\vx}$ shares similar features as $\vx$ in $\sD$ while $\tilde{y}$ could be random.

\vspace{-1mm}
\section{Gradient embedding perturbation}
\vspace{-1mm}

 An overview of GEP is given in Figure~\ref{fig:overview}. GEP has three major ingredients: 1) first, estimate an anchor subspace that contains the principal components of some non-sensitive anchor gradients via power method; 2) then, project private gradients into the anchor subspace and produce low-dimensional embeddings of private gradients and residual gradients; 3) finally, perturb gradient embedding and residual gradient separately to establish differential privacy guarantee. In Section~\ref{sec:ppg_algo}, we present the GEP algorithm in detail.  In Section~\ref{sec:ppgr_error}, we given an analysis on the residual gradients. In Section~\ref{sec:learning_ppg}, we give a differentially private learning algorithm that updates the model with the output of GEP.

\vspace{-1mm}
\subsection{The GEP algorithm and its privacy analysis}
\label{sec:ppg_algo}
\vspace{-1mm}

The pseudocode of GEP is presented in Algorithm~\ref{alg:ppg}. For convenience, we write a set of gradients and a set of basis vectors as matrices with each row being one gradient/basis vector. 

The anchor subspace is constructed as follows.   We first compute the gradients of the model on an auxiliary dataset $\sD^{(a)}$ with $m$ samples, which is referred to as the anchor gradients $\mG^{(a)}\in \sR^{m\times p}$. We then use the power method to estimate the principal components of $\mG^{(a)}$ to construct a subspace basis $\mB\in \sR^{k\times p}$, which is referred to as the anchor subspace. All these matrices are publishable  because $\sD^{(a)}$ is non-sensitive. We expect that the anchor subspace $\mB$ can cover most energy of private gradients when the auxiliary data are not far from private data and $m,k$ are reasonably large.

Suppose that the private gradients are $\mG\in \sR^{n\times p}$. Then, we project the private gradients into the anchor subspace $\mB$.  The projection  produces low-dimensional embeddings $\mW = \mG \mB^T$ and residual gradients $\mR =\mG - \mG \mB^T\mB$. The magnitude of residual gradients is usually much smaller than original gradient even when $k$ is small because of the  gradient redundancy.

Then, we aggregate the gradient embeddings and the residual gradients, respectively. We perturb the aggregated  embedding and the aggregated residual gradient respectively to guarantee certain differential privacy. Finally, we release the perturbed embedding and the perturbed residual gradient and construct an unbiased estimator of the private gradient: $\tilde{\vv} := (\tilde \vw^{T}\mB+\tilde \vr)/n$.  This construction process does not resulting in additional privacy loss because of DP's post-processing property. The privacy analysis of the whole process of GEP is given in Theorem~\ref{thm:privacy_ppgr}.

\begin{algorithm}
\caption{Gradient embedding perturbation}
   \label{alg:ppg}
\begin{algorithmic} [1]
   \STATE {\bfseries Input:} anchor gradients $\mG^{(a)}\in\mathbb{R}^{m\times p}$; number of basis vectors $k$; private gradients $\mG\in\mathbb{R}^{n\times p}$; clipping thresholds $S_{1},S_{2}$; standard deviations $\sigma_{1}, \sigma_{2}$; number of power iterations $t$.

   \medskip
   \STATE //\textsl{First stage: Compute an orthonormal  basis for the anchor subspace.}
   \STATE Initialize $\mB\in\mathbb{R}^{k\times p}$ randomly.
   
   \FOR{$i=1$ {\bfseries to} $t$}
			\STATE Compute $\mA=\mG^{(a)}\mB^{T}$ and $\mB=\mA^{T}\mG^{(a)}$.
			\STATE Orthogonalize $\mB$ and normalize row vectors.
   \ENDFOR
   
   \STATE Delete $\mG^{(a)}$ to free memory.
   
   \medskip
   \STATE //\textsl{Second stage: project the private gradients $\mG$ into anchor subspace $\mB$}
   \STATE Compute gradient embeddings $\mW=\mG\mB^{T}$ and clip its rows with $S_{1}$ to obtain $\hat \mW$.
   
   \STATE Compute residual gradients $\mR=\mG-\mW\mB$ and clip its rows with $S_{2}$ to obtain $\hat \mR$.
   
   \medskip
   \STATE //\textsl{Third stage: perturb gradient embedding and residual gradient separately}
   \STATE Perturb embedding with noise $\vz^{(1)} \sim \mathcal{N}(0, \sigma_1^2 \mI_{k\times k})$: $\;\vw := \sum_{i}\hat\mW_{i,:},\;\;  \tilde \vw := \vw+\vz^{(1)}.$
   
   \STATE Perturb residual gradient with noise $\vz^{(2)} \sim \mathcal{N}(0, \sigma_2^2\mI_{p\times p})$:
   $\;\vr:=\sum_{i} \hat\mR_{i,:}, \;\; \tilde \vr :=  \vr + \vz^{(2)}.$
   
   \STATE Return $\tilde{\vv} := (\tilde \vw^{T}\mB+\tilde \vr)/n$.

\end{algorithmic}
\end{algorithm}

\begin{restatable}{thm}{privacyppgr}
\label{thm:privacy_ppgr}
Let $S_{1}$ and $S_{2}$ be the sensitivity of $\vw$ and $\vr$, respectively, the output of Algorithm~\ref{alg:ppg} satisfies $(\epsilon,\delta)$-DP for any $\delta\in (0,1)$ and $\epsilon\leq 2\log(1/\delta)$   if we choose $\sigma_{1}\geq  2S_{1}\sqrt{2\log(1/\delta)}/\epsilon$ and $\sigma_{2}\geq  2S_{2}\sqrt{2\log(1/\delta)}/\epsilon$.
\end{restatable}

A common practice to control sensitivity is to clip the output with a pre-defined threshold. In our experiments, we use different thresholds $S_{1}$ and $S_{2}$ to clip the gradient embeddings and residual gradients, respectively. The privacy loss of GEP consists of two parts: the privacy loss incurred by releasing the perturbed embedding and the privacy loss incurred by releasing the perturbed residual gradient. We compose these two parts via the Rényi differential privacy and convert it to $(\epsilon,\delta)$-DP.

We highlight several implementation techniques that make GEP widely applicable and implementable with reasonable computational cost.  Firstly,  auxiliary non-sensitive data do not have to be the same source as the private data and the auxiliary data can be randomly labeled. This  non-sensitive data assumption is very weak and easy to satisfy in practical scenarios. To understand why random label works, a quick example is that for the least squares regression problem the individual gradient is aligned with the feature vector while the label only scales the length but does not change the direction. This auxiliary data assumption avoids conducting principal component analysis (PCA) on private gradients, which requires releasing private high-dimensional basis vectors and hence introduces large privacy loss. Secondly,  we use \emph{power method} \citep{panju2011iterative, vogels2019powersgd} to approximately estimate the principal components.  The  new operation we introduce is standard matrix multiplication that enjoys efficient implementation on GPU. The computational complexity of each power iteration is $2mkp$,  where $p$  is the number of model parameters, $m$ is the number of anchor gradients and $k$ is the number of subspace basis vectors. Thirdly, we divide the parameters into different groups and compute one orthonormal basis for each group. This further reduces the computational cost. For example, suppose the parameters are divided into two groups with size $p_{1},p_{2}$ and the numbers of basis vectors are $k_{1}, k_{2}$, the computational complexity of each power iteration is $2m(k_{1}p_{1}+k_{2}p_{2})$, which is smaller than $2m(k_{1}+k_{2})(p_{1}+p_{2})$. In Appendix~\ref{sec:complexity}, we analyze the additional computational and memory costs of GEP compared to standard gradient perturbation.

Curious readers may wonder if we can use random projection to reduce the dimensionality as Johnson–Lindenstrauss Lemma \citep{dasgupta2003elementary}  guarantees that one can preserve the pairwise distance between any two points after projecting into a random subspace of much lower dimension. However, preserving the pairwise distance is not sufficient for high quality gradient reconstruction, which is verified by the empirical observation in Appendix~\ref{apdx:abla}.

\vspace{-1mm}
\subsection{An analysis on the residual gradients of GEP}
\label{sec:ppgr_error}
\vspace{-1mm}

Let $\vg :=\frac{1}{n} \sum_{i}\mG_{i,:}$ be the target private gradient. For a given anchor subspace $\mB$, the residual gradients are defined as $\mR:=\mG-\mG\mB^{T}\mB$. We then analyze how large the residual gradients could be. The following argument holds for all time steps and we ignore the time step index for simplicity.

For the ease of discussion, we introduce $\vxi_i := (\mG_{i,:})^T$ for $i \in [n]$ to denote the the private gradients and the $\hat{\vxi}_j := (\mG^{(a)}_{j,:})^T$ for $j \in [m]$  to denote the anchor gradients. We use $\lambda_{k}(\cdot)$ to denote the $k_{th}$ largest eigenvalue of a given matrix. We assume that the private gradients $\vxi_1, ..., \vxi_n$ and the anchor gradients $\hat{\vxi}_1, ..., \hat{\vxi}_m$ are sampled independently from a distribution $\gP$. We assume $\mSigma := \E_{\vxi\sim \gP} \vxi \vxi^T \in \sR^{p\times p}$ to be the population gradient (uncentered) covariance matrix. We also consider the (uncentered) empirical gradient covariance matrix $\hat{\mS} := \frac{1}{m}\sum_{i=1}^m\hat{\vxi}_i\hat{\vxi}_i^T$.

One case is that the population gradient covariance matrix $\mSigma$ is low-rank $k$. In this case we can argue that the residual gradients are $0$ once the number of anchor gradients $m>k$.

\begin{restatable}{lemma}{lowrankerror}
\label{lma:lowrank_error}
Assume that the population covariance matrix $\mSigma$ is with rank $k$ and the distribution $\gP$ satisfies $\sP(\vxi \in \sF_s) =0$ for all $s$-flats $\sF_s$ in $\sR^p$ with $0\le s < k$. Let $\mSigma = \mV_k \Lambda \mV_k^T$  and $\hat{\mS} = \hat{\mV}_{k'} \hat{\Lambda} \hat{\mV}_{k'}^T$ be the eigendecompositions of $\mSigma$ and the empirical covariance matrix $\hat{\mS}$, respectively, such that $\lambda_{k'}(\hat{\mS}) >0$ and $\lambda_{k'+1}(\hat{\mS})=0$.  Then if $m\ge k$, we have with probability 1,
\begin{flalign}
k'= k \quad \text{and} \quad \|\mV_k\mV_k^T - \hat{\mV}_k \hat{\mV}_k^T\|_2 = 0.
\end{flalign}
\end{restatable}
\begin{proof}
The proof is based on the non-singularity of covariance matrix. See Appendix \ref{sec:proof_dp}. 
\end{proof}

We note that  $s$-flat is the translate $\sF_s = {\vx} + \sF_{s(0)}$ of an $s$-dimensional linear subspace $\sF_{s(0)}$ in $\sR^p$ and the normal distribution satisfies such condition \citep{eaton1973non,muirhead2009aspects}. Therefore, we have seen that for low-rank case of population covariance matrix, the residual gradients are 0 once $m>k$. In the general case, we measure the expected norm of the residual gradients.

\begin{lemma}
\label{lma:approx_error}
Assume that $\vxi \sim \gP$ and $\|\vxi\|^2<T$ almost surely. Let $\mSigma = \mV \Lambda \mV^T$ be the eigendecomposition of the population covariance matrix $\mSigma$. Let $\hat{\mS} = \hat{\mV}_k \hat{\Lambda} \hat{\mV}_k^T$ be the eigendecomposition of the empirical covariance matrix $\hat{\mS}$. Then we have with probability $1-2\exp(-\delta)$,
\begin{flalign}
\E\|\vxi - \Pi_{\hat{\mV}_k}(\vxi)\|_2^2 \le \sum_{k'>k} \lambda_{k'}(\mSigma) + \sqrt{kC/m} + T \sqrt{2\delta/m},
\end{flalign}
where $C= \left[\E \|\vxi\|^4 - \sum_i \lambda_i^2(\mSigma)\right] + \left[\frac{1}{m}\sum_{j=1}^m \|\hat\vxi_j\|^4 - \sum_i \lambda_i^2(\hat\mS)\right]$, $\Pi_{\hat{\mV}_k}$ is a projection operator onto the subspace $\hat{\mV}_k$ and the $\E$ is taken over the randomness of $\vxi\sim \gP$.
\end{lemma}

\begin{proof}
The proof is an adaptation of Theorem 3.1 in \cite{blanchard2007statistical}.
\end{proof}

From Lemma~\ref{lma:approx_error}, we can see the larger the number of anchor gradients and the dimension of the anchor subspace $k$, the smaller the residual gradients. We can choose $m,k$ properly such that the upper bound on the expected residual gradient norm is small. This indicates that we may use a smaller clipping threshold and consequently apply smaller noises with achieving the same privacy guarantee. 

\begin{figure}
\centering

    \begin{minipage}{0.64\textwidth}
        \begin{center}
        \centerline{\includegraphics[width=0.8\linewidth]{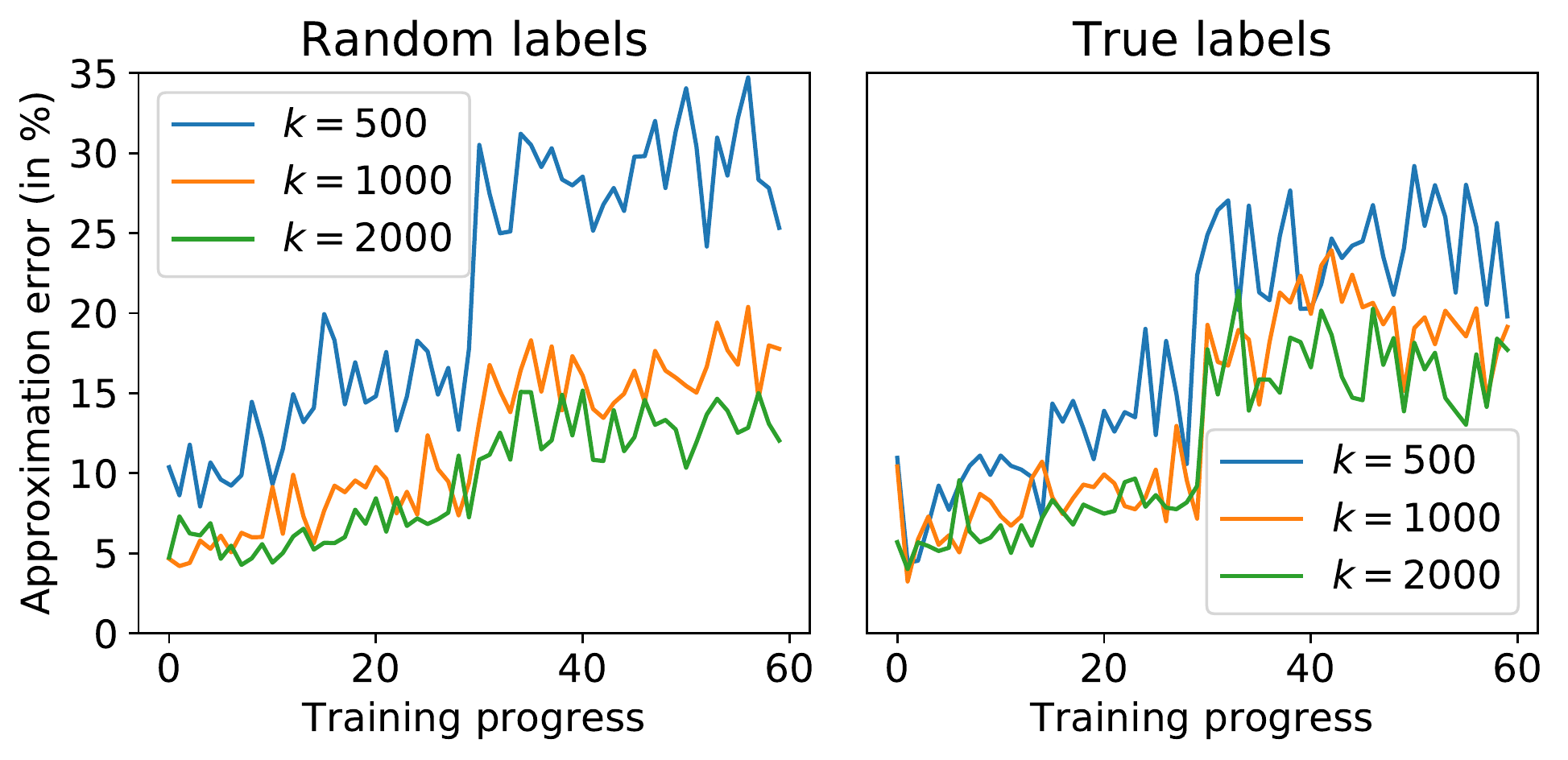}}
          \caption{ Relative projection error  ($\norm{\frac{1}{n}\vr}/\norm{\vg}$) of the second stage in ResNet20. The number of anchor gradients is $2000$. The dimension of anchor subspace is $k$. The learning rate is decayed by $10$ at epoch 30. The left plot uses random samples from ImageNet. The right plot uses random samples from test data. The benefit of increasing $k$ becomes smaller when $k$ is larger. }
          \label{fig:approx_error}
        \end{center}
    \end{minipage}
    \hfill
    \begin{minipage}{0.33\textwidth}
        \begin{center}
        \centerline{\includegraphics[width=0.8\linewidth]{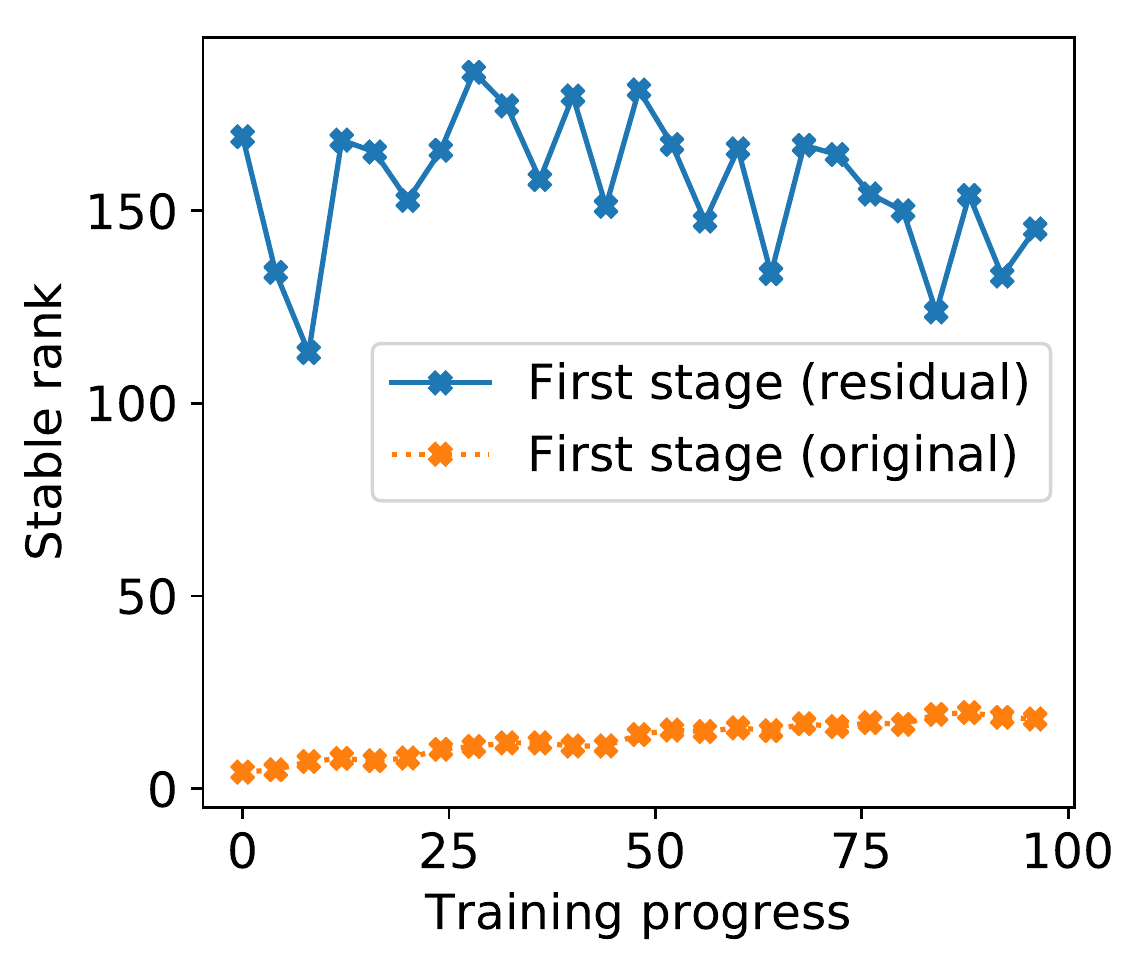}}
        \caption{Stable rank of the residual gradient matrix versus original gradient matrix. The gradients are computed on full batch data for  the first stage in ResNet20. The dimension of anchor subspace is $k=1000$.   } 
        \label{fig:rgp_stable_rank}
        \end{center}
    \end{minipage}    
\end{figure}

We next empirically examine the projection error $\vr=\sum_{i}\mR_{i,:}$ by training a $20$-layer ResNet on CIFAR10 dataset. We try two different types of auxiliary data to compute the anchor gradients: 1) samples from the same source as private data with correct labels, i.e., $2000$ random samples from the test data;  2) samples from different source with random labels, i.e., $2000$ random samples from ImageNet. The relation between the dimension of anchor subspace $k$  and the projection error rate ($\norm{\frac{1}{n}\vr}/\norm{\vg}$) is presented in Figure~\ref{fig:approx_error}. We can see that the project error is small and decreases with $k$, and the benefit of  increasing $k$ diminishes when $k$ is large, which is implied by Lemma \ref{lma:approx_error}. In practice one can  only use small or moderate $k$ because of the memory constraint. GEP needs to store at least $k$ individual gradients and each individual gradient consumes the same amount of memory as the model itself.  Moreover, we can see that the projection into anchor subspace of random labeled auxiliary data yields comparable  projection error, corroborating our argument that unlabeled auxiliary data are sufficient for finding the anchor subspace. 

We also verify that the redundancy of residual gradients is small, by plotting the stable rank  of residual gradient matrix in Figure~\ref{fig:rgp_stable_rank}. The stable rank of residual gradient matrix is an order of magnitude higher than the stable rank of original gradient matrix. This implies that it could be hard to further approximate $\mR$ with low-dimensional embeddings.

We next compare the GEP with a scheme that simply discards the residual gradients and only outputs the perturbed gradient embedding, i.e., the output is $\tilde{\vu} := \tilde \vw^{T}\mB/n$.

\begin{remark}
\label{thm:error_cp}
Let $\tilde \vu := \tilde \vw^{T}\mB/n$ be the reconstructed gradient from noisy gradient embedding and $\tilde{\vv}$ be the output of GEP. If ignoring the effect of gradient clipping, we have 
\begin{flalign}
\mathbb{E}[\tvu]=  \vg-\vr/n, \quad \E [\tvv] = \vg. 
\end{flalign}
        
where $\vr=\sum_{i}\mR_{i,:}$ is the aggregated residual gradients, $\tilde\vw, \mB$  are given in Algorithm~\ref{alg:ppg} and the expectation is over the added random noises.
\end{remark}

This indicates that $\tvu$ contains a systematic error that makes $\tvu$ always deviate from  $\vg$ by the residual gradient. This  systematic error is the projection error, which is plotted in Figure~\ref{fig:approx_error}. The systematic error cannot be mitigated by reducing the noise magnitude (e.g., increasing the privacy budget or collecting more private data).  We refer to the algorithm releasing $\tvu$ directly as \emph{Biased-GEP} or \emph{B-GEP} for short, which can be viewed as an efficient implementation of the algorithm in \citep{zhou2020bypassing}. In our experiments, B-GEP can outperform standard gradient perturbation when $k$ is large but is  inferior to GEP. We note that the above remark is made with ignoring the clipping effect (or set a large clipping threshold). In practice, we do apply clipping for the individual gradients at each time step, which makes the expectations in Remark \ref{thm:error_cp} obscure \citep{chen2020understanding}. We note that the claim that $\tvv$ is an unbiased estimator of $\vg$ is not that precise when applying gradient clipping.

\vspace{-1mm}
\subsection{Private learning with gradient embedding perturbation}
\vspace{-1mm}
\label{sec:learning_ppg}

GEP (Algorithm \ref{alg:ppg}) describes how to release one-step gradient with privacy guarantee. In this section, we compose the privacy losses at each step to establish the privacy guarantee for the whole learning process. The differentially private learning process with GEP is given in Algorithm~\ref{alg:dp_gd} and the privacy analysis is presented in Theorem~\ref{thm:privacy_gd}.

\begin{algorithm}
\caption{Differentially private gradient descent with GEP.}
  \label{alg:dp_gd}
\begin{algorithmic} [1]
  \STATE {\bfseries Input:} private dataset $\sD$; auxiliary dataset $\sD^{(a)}$;  number of updates $T$; learning rate $\eta$; configuration of GEP $\sC$; loss function $\ell$;

  \STATE {\bfseries Output:} Differentially private model $\boldsymbol\theta_{T}$. 
   
  %\medskip
 
  \FOR{$t=0$ {\bfseries to} $T-1$}
            \STATE  Compute the private gradients $\mG_{t}$ and anchor gradients $\mG^{(a)}_{t}$ of loss with respect to  $\boldsymbol\theta_{t}$.
		    \STATE  Call GEP   with $\mG_{t},\mG^{(a)}_{t}$ and configuration $\sC$ to get  $\tilde \vv_{t}$.
			\STATE  Update model $\boldsymbol\theta_{t+1}=\boldsymbol\theta_{t}-\eta\tilde\vv_{t}$.
  \ENDFOR
\end{algorithmic}
\end{algorithm}

\begin{restatable}{thm}{privacygd}
\label{thm:privacy_gd}
 For any $\epsilon<2\log(1/\delta)$ and $\delta\in(0,1)$, the output of Algorithm~\ref{alg:dp_gd} satisfies $(\epsilon,\delta)$-DP if we set  $\sigma\geq 2\sqrt{2T\log(1/\delta)}/\epsilon$.
\end{restatable}

If the private gradients are randomly sampled from the full batch gradients, the privacy guarantee can be strengthened via the \emph{privacy amplification by subsampling} theorem of DP \citep{balle2018privacy,wang2019subsampled,zhu2019poission, mironov2019renyi}.  Theorem~\ref{thm:convergence} gives the expected excess error of Algorithm~\ref{alg:dp_gd}. Expected excess error measures the distance between the algorithm's output and the optimal solution in expectation.

\begin{restatable}{thm}{convergence}
\label{thm:convergence}
Suppose the loss $L(\boldsymbol\theta)=\frac{1}{n}\sum_{(\vx,y)\in\sD}\ell(f_{\boldsymbol\theta}(\vx),y)$ is 1-Lipschitz, convex, and $\beta$-smooth. If $\eta=\frac{1}{\beta}$, $T=\frac{n\beta\epsilon}{\sqrt{p}}$, and $\boldsymbol{\bar \theta}=\frac{1}{T}\sum_{t=1}^{T}\boldsymbol\theta_{t}$, then we have 
$\mathbb{E}[L(\boldsymbol{\bar\theta})]-L(\boldsymbol{\theta_{*}})\leq \mathcal{O}\left(\frac{\sqrt{k\log(1/\delta)}}{n\epsilon}+\frac{\bar r\sqrt{p\log(1/\delta)}}{n\epsilon}\right)$,
where $\bar r=\frac{1}{T}\sum_{t=0}^{T-1}r^{2}_{t}$ and $r_{t}=\max_{i} \norm{(\mR_{t})_{i,:}}$ is the sensitivity of  residual gradient at step $t$. 

\end{restatable}

 The $\bar r$ term represents the average projection error over the training process. The previous best expected excess error for gradient perturbation is $\mathcal{O}(\sqrt{p\log(1/\delta)}/(n\epsilon))$ \citep{wang2017differentially}. As shown in Lemma \ref{lma:lowrank_error}, if the gradients locate in a $k$-dimensional subspace over the training process, $\bar{r}=0$ and the excess error is $\mathcal{O}(\sqrt{k\log(1/\delta)}/(n\epsilon))$, independent of the problem ambient dimension $p$. When the gradients are in general position, i.e., gradient matrix is not exact low-rank, Lemma \ref{lma:approx_error} and the empirical result give a hint on how small the residual gradients could be. However, it is hard to get a good bound on $\max_i\|(\mR_t)_{i,:}\|$ and the bound in Theorem \ref{thm:convergence} does not explicitly improve over previous result. One possible solution is to use a clipping threshold based on the expected residual gradient norm. Then the output gradient becomes biased because of clipping and the utility/privacy guarantees  in Theorem \ref{thm:convergence}/\ref{thm:privacy_gd} require new elaborate derivation. We leave this for future work.
\vspace{-2mm}
\section{Experiments}
\label{sec:exp}
\vspace{-2mm}

We conduct experiments on  MNIST, extended SVHN, and CIFAR-10 datasets. Our implementation is publicly available\footnote{\url{https://github.com/dayu11/Gradient-Embedding-Perturbation}}. The model for MNIST has two convolutional layers with max-pooling and one fully connected layer.  The model for SVHN and CIFAR-10 is ResNet20 in \cite{he2016deep}. We replace all batch normalization  \citep{ioffe2015batch} layers with group normalization \citep{wu2018group} layers because batch normalization mixes the representations of different samples and makes  the privacy loss cannot be analyzed accurately. The non-private accuracy for MNIST, SVHN, and CIFAR-10 is 99.1\%, 95.9\%, and 90.4\%, respectively. 

We also provide experiments with  pre-trained models in Appendix~\ref{sec:linear}. \cite{tramer2020differentially} show that differentially private linear classifier can achieve high accuracy using the features produced by pre-trained models. We examine whether GEP can improve the performance of such private linear classifiers.  Notably, using the features produced by a model pre-trained on unlabeled ImageNet, GEP achieves 94.8\% validation accuracy on CIFAR10 with  $\epsilon=2$.

\textbf{Evaluated algorithms} We use the algorithm in \cite{abadi2016deep} as benchmark gradient perturbation approach, referred to as ``GP''. We also compare GEP with PATE \citep{papernot2016semi}. We run the experiments for PATE using the official implementation.  The privacy parameter $\epsilon$ of PATE is  data-dependent and hence cannot be released directly (see Section 3.3 in \cite{papernot2016semi}). Nonetheless, we report the results of PATE anyway.

\textbf{Implementation details}     At each step, GEP needs to release two vectors: the noisy gradient embedding and the  noisy residual gradient. The gradient embeddings have a sensitivity of $S_{1}$ and the residual gradients have a sensitivity of $S_{2}$ because of the clipping. The output of GEP can be constructed as follows: (1) normalize the gradient embeddings and residual gradients by $1/S_{1}$ and $1/S_{2}$, respectively, (2) concatenate the rescaled vectors, (3) release the concatenated vector via gaussian mechanism with sensitivity $\sqrt{2}$, (4) rescale the two components by $S_{1}$ and $S_{2}$.  B-GEP only needs to release the normalized noisy gradient embedding. We  use the numerical tool in \cite{mironov2019renyi} to compute the privacy loss. For given privacy budget and sampling probability, $\sigma$ is set to be the smallest value such that the privacy budget is allowable to run desired epochs. 

All experiments are run on a single Tesla V100 GPU with 16G memory. For ResNet20, the parameters are divided into five groups: input layer, output layer, and three intermediate stages. For a given quota of basis vectors, we allocate it to each group according to the square root of  the number of parameters in each group.  We compute an orthonormal subspace basis  on each group separately.  Then we concatenate the projections of all groups to construct gradient embeddings.  The number of power iterations $t$ is set as $1$ as empirical evaluations suggest more iterations do not improve the performance for GEP and B-GEP.

 For all datasets, the anchor gradients are computed on $2000$  random samples from ImageNet.  In Appendix~\ref{apdx:abla}, we examine the influence of choosing different numbers of anchor gradients and different sources of auxiliary data. The selected images are downsampled into size of $32\times 32$ ($28\times 28$ for MNIST) and  we  label them randomly at each update.  For SVHN and CIFAR-10,  $k$ is chosen from $[500, 1000, 1500, 2000]$. For MNIST, we halve the size of $k$. We use SGD with momentum  0.9 as the  optimizer.  Initial learning rate and batchsize are $0.1$ and $1000$, respectively. The learning rate is divided by $10$ at middle of training. Weight decay is set as $1\times 10^{-4}$. The clipping threshold for is $10$ for original gradients and $2$ for residual gradients.  The number of training epochs for CIFAR-10 and MNIST is 50, 100, 200 for privacy parameter $\epsilon=2,5,8$, respectively. The number of training epochs for SVHN is 5, 10, 20 for privacy parameter $\epsilon=2,5,8$, respectively.     Privacy parameter $\delta$ is $1\times 10^{-6}$ for SVHN and $1\times 10^{-5}$ for CIFAR-10 and MNIST.

\textbf{Results} The best accuracy with given $\epsilon$ is in Table~\ref{tbl:no_fine-tune_result}.   For all datasets, GEP achieves considerable improvement over GP in \cite{abadi2016deep}.   Specifically, GEP achieves $74.9\%$ test accuracy on CIFAR-10 with $(8,10^{-5})$-DP, outperforming GP by  $18.5\%$. PATE achieves best accuracy on MNIST  
but its performance drops  as the dataset becomes more complex.  

\begin{figure}
    \centering
  \includegraphics[width=0.75\linewidth]{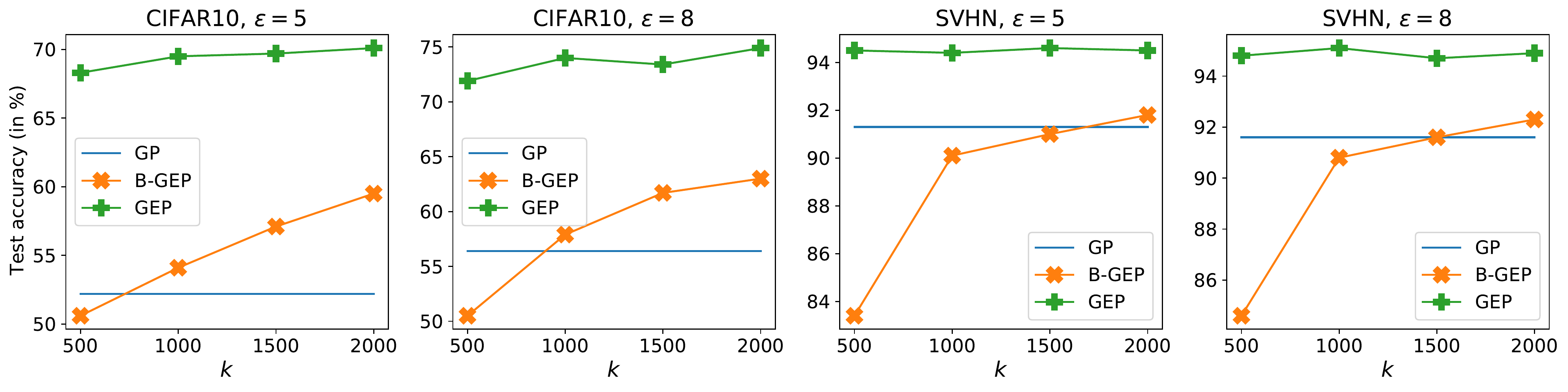}
  \caption{Test accuracy when varying the dimension of anchor subspace. GEP significantly outperforms B-GEP for all $k$. Moreover, the performance of GEP is not that sensitive to $k$.   }
  \label{fig:influence_k}
\end{figure}

We also plot the relation between accuracy and $k$ in Figure~\ref{fig:influence_k}.  GEP is less sensitive to the choice of $k$ and outperforms B-GEP for all choices of $k$. The improvement of increasing $k$ becomes smaller as $k$ becomes larger. We note that the memory cost of choosing large $k$ is high because we need to store at least $k$ individual gradients to compute anchor subspace.

\begin{table}
\label{tbl:no_fine-tune_result}
\small
\centering
    \caption{Test accuracy (in \%) with varying choices of privacy bound $\epsilon$. The numbers under symbol $\Delta$ denote the improvement over GP baseline. }
    
        \begin{tabular}{ll|l|l|l|l|l|l}
        \hline
        \hline
            Dataset                   &    Algorithm          & $\epsilon=2$  &  $\Delta$            & $\epsilon=5$  &  $\Delta$  &  $\epsilon=8$ & $\Delta$  \\
            \hline
            \multirow{4}{*}{MNIST}  &  GP &      94.7       &       +0.0       &   96.8    &  +0.0  &      97.2    & +0.0 \\\cline{2-8}
                                 &  PATE &       98.5       &     \textbf{+3.8}     &  98.5      &    \textbf{+1.7}   &     98.6     &  \textbf{+1.4}    \\\cline{2-8}
                                  &  B-GEP  &        93.1     &   -1.6        &   94.5  &   -2.3  &    95.9      &  -1.3  \\\cline{2-8}
                                   &  GEP  &      96.3      & +1.6     &  97.9   & \textbf{+1.1}   &      98.4   &   \textbf{+1.2}    \\\hline \hline
            \multirow{4}{*}{SVHN}  &   GP &       87.1     &       +0.0       &    91.3  &  +0.0  &  91.6       & +0.0 \\\cline{2-8}
                                 &  PATE &       80.7      & -6.4    &    91.6   & +0.3     &     91.6     &  +0.0    \\\cline{2-8}
                                  &  B-GEP  &      88.5       & +1.4 &   91.8 &  +0.5   &      92.3    &  +0.7 \\\cline{2-8}
                                   &  GEP  &          92.3   &   \textbf{+5.2}   &   94.7  & \textbf{+3.4}  &      95.1   &   \textbf{+3.5}      \\\hline \hline
            \multirow{3}{*}{CIFAR-10}  &  GP &      43.6       & +0.0  &     52.2   &  +0.0   & 56.4\tablefootnote{The test accuracy of DP-SGD can be improved to $\sim$ 62\% by tuning the hyperparameters. See the implementation in \url{https://github.com/dayu11/Differentially-Private-Deep-Learning}.} & +0.0 \\\cline{2-8}
                                    &  PATE &       34.2      &  -9.4  &     41.9   &   -10.3   &    43.6     &    -12.8  \\\cline{2-8}
                                    &  B-GEP &      50.3       &  +6.7  &    59.5    &    +7.3      &   63.0          & +6.6  \\\cline{2-8}
                                     &  GEP &        59.7     & \textbf{+16.1}  &   70.1   &  \textbf{+17.9} &     74.9   &     \textbf{+18.5}  \\\hline
                                        \hline
        \end{tabular}
\end{table}

\vspace{-2mm}
\section{Conclusion}
\vspace{-2mm}

In this paper, we propose  Gradient Embedding Perturbation (GEP) for learning with differential privacy. GEP leverages the gradient redundancy to reduce the added noise and outputs an unbiased estimator of target gradient. The several key designs of GEP significantly boost the applicability of GEP.  Extensive experiments on real world datasets demonstrate the superior utility of GEP.

%\newpage

\bibliography{privacy}

\begin{thebibliography}{61}
\providecommand{\natexlab}[1]{#1}
\providecommand{\url}[1]{\texttt{#1}}
\expandafter\ifx\csname urlstyle\endcsname\relax
  \providecommand{\doi}[1]{doi: #1}\else
  \providecommand{\doi}{doi: \begingroup \urlstyle{rm}\Url}\fi

\bibitem[Abadi et~al.(2016)Abadi, Chu, Goodfellow, McMahan, Mironov, Talwar,
  and Zhang]{abadi2016deep}
Martin Abadi, Andy Chu, Ian Goodfellow, H~Brendan McMahan, Ilya Mironov, Kunal
  Talwar, and Li~Zhang.
\newblock Deep learning with differential privacy.
\newblock In \emph{ACM SIGSAC Conference on Computer and Communications
  Security}, 2016.

\bibitem[Alon et~al.(2019)Alon, Bassily, and Moran]{alon2019limits}
Noga Alon, Raef Bassily, and Shay Moran.
\newblock Limits of private learning with access to public data.
\newblock In \emph{Advances in Neural Information Processing Systems}, 2019.

\bibitem[Balle et~al.(2018)Balle, Barthe, and Gaboardi]{balle2018privacy}
Borja Balle, Gilles Barthe, and Marco Gaboardi.
\newblock Privacy amplification by subsampling: Tight analyses via couplings
  and divergences.
\newblock In \emph{Advances in Neural Information Processing Systems}, 2018.

\bibitem[Bassily et~al.(2014)Bassily, Smith, and
  Thakurta]{bassily2014differentially}
Raef Bassily, Adam Smith, and Abhradeep Thakurta.
\newblock Differentially private empirical risk minimization: Efficient
  algorithms and tight error bounds.
\newblock \emph{Annual Symposium on Foundations of Computer Science}, 2014.

\bibitem[Bernau et~al.(2019)Bernau, Grassal, Robl, and
  Kerschbaum]{bernau2019assessing}
Daniel Bernau, Philip-William Grassal, Jonas Robl, and Florian Kerschbaum.
\newblock Assessing differentially private deep learning with membership
  inference.
\newblock \emph{arXiv preprint arXiv:1912.11328}, 2019.

\bibitem[Blanchard et~al.(2007)Blanchard, Bousquet, and
  Zwald]{blanchard2007statistical}
Gilles Blanchard, Olivier Bousquet, and Laurent Zwald.
\newblock Statistical properties of kernel principal component analysis.
\newblock \emph{Machine Learning}, 66\penalty0 (2-3):\penalty0 259--294, 2007.

\bibitem[Bun \& Steinke(2016)Bun and Steinke]{bun2016concentrated}
Mark Bun and Thomas Steinke.
\newblock Concentrated differential privacy: Simplifications, extensions, and
  lower bounds.
\newblock In \emph{Theory of Cryptography Conference}, 2016.

\bibitem[Carlini et~al.(2019)Carlini, Liu, Erlingsson, Kos, and
  Song]{carlini2019secret}
Nicholas Carlini, Chang Liu, {\'U}lfar Erlingsson, Jernej Kos, and Dawn Song.
\newblock The secret sharer: Evaluating and testing unintended memorization in
  neural networks.
\newblock In \emph{USENIX Security Symposium}, 2019.

\bibitem[Chen et~al.(2020{\natexlab{a}})Chen, Kornblith, Swersky, Norouzi, and
  Hinton]{chen2020big}
Ting Chen, Simon Kornblith, Kevin Swersky, Mohammad Norouzi, and Geoffrey
  Hinton.
\newblock Big self-supervised models are strong semi-supervised learners.
\newblock \emph{arXiv preprint arXiv:2006.10029}, 2020{\natexlab{a}}.

\bibitem[Chen et~al.(2020{\natexlab{b}})Chen, Wu, and
  Hong]{chen2020understanding}
Xiangyi Chen, Steven~Z Wu, and Mingyi Hong.
\newblock Understanding gradient clipping in private sgd: A geometric
  perspective.
\newblock \emph{Advances in Neural Information Processing Systems}, 33,
  2020{\natexlab{b}}.

\bibitem[Dasgupta \& Gupta(2003)Dasgupta and Gupta]{dasgupta2003elementary}
Sanjoy Dasgupta and Anupam Gupta.
\newblock An elementary proof of a theorem of johnson and lindenstrauss.
\newblock \emph{Random Structures \& Algorithms}, 2003.

\bibitem[Dong et~al.(2019)Dong, Roth, and Su]{dong2019gaussian}
Jinshuo Dong, Aaron Roth, and Weijie~J Su.
\newblock Gaussian differential privacy.
\newblock \emph{arXiv preprint arXiv:1905.02383}, 2019.

\bibitem[Dwork et~al.(2006{\natexlab{a}})Dwork, Kenthapadi, McSherry, Mironov,
  and Naor]{dwork2006our}
Cynthia Dwork, Krishnaram Kenthapadi, Frank McSherry, Ilya Mironov, and Moni
  Naor.
\newblock Our data, ourselves: Privacy via distributed noise generation.
\newblock In \emph{Annual International Conference on the Theory and
  Applications of Cryptographic Techniques}, 2006{\natexlab{a}}.

\bibitem[Dwork et~al.(2006{\natexlab{b}})Dwork, McSherry, Nissim, and
  Smith]{dwork2006calibrating}
Cynthia Dwork, Frank McSherry, Kobbi Nissim, and Adam Smith.
\newblock Calibrating noise to sensitivity in private data analysis.
\newblock In \emph{Theory of cryptography conference}, 2006{\natexlab{b}}.

\bibitem[Dwork et~al.(2014)Dwork, Roth, et~al.]{algofound}
Cynthia Dwork, Aaron Roth, et~al.
\newblock The algorithmic foundations of differential privacy.
\newblock \emph{Foundations and Trends{\textregistered} in Theoretical Computer
  Science}, 2014.

\bibitem[Eaton \& Perlman(1973)Eaton and Perlman]{eaton1973non}
Morris~L Eaton and Michael~D Perlman.
\newblock The non-singularity of generalized sample covariance matrices.
\newblock \emph{The Annals of Statistics}, pp.\  710--717, 1973.

\bibitem[Fredrikson et~al.(2015)Fredrikson, Jha, and
  Ristenpart]{fredrikson2015model}
Matt Fredrikson, Somesh Jha, and Thomas Ristenpart.
\newblock Model inversion attacks that exploit confidence information and basic
  countermeasures.
\newblock In \emph{ACM SIGSAC Conference on Computer and Communications
  Security}, 2015.

\bibitem[Fukuchi et~al.(2017)Fukuchi, Tran, and
  Sakuma]{fukuchi2017differentially}
Kazuto Fukuchi, Quang~Khai Tran, and Jun Sakuma.
\newblock Differentially private empirical risk minimization with input
  perturbation.
\newblock In \emph{International Conference on Discovery Science}, 2017.

\bibitem[Gooneratne et~al.(2020)Gooneratne, Sim, Zadrazil, Kabel, Beaufays, and
  Motta]{gooneratne2020low}
Mary Gooneratne, Khe~Chai Sim, Petr Zadrazil, Andreas Kabel, Fran{\c{c}}oise
  Beaufays, and Giovanni Motta.
\newblock Low-rank gradient approximation for memory-efficient on-device
  training of deep neural network.
\newblock In \emph{IEEE International Conference on Acoustics, Speech and
  Signal Processing (ICASSP)}. IEEE, 2020.

\bibitem[Gur-Ari et~al.(2018)Gur-Ari, Roberts, and Dyer]{gur2018gradient}
Guy Gur-Ari, Daniel~A Roberts, and Ethan Dyer.
\newblock Gradient descent happens in a tiny subspace.
\newblock \emph{arXiv preprint arXiv:1812.04754}, 2018.

\bibitem[He et~al.(2016)He, Zhang, Ren, and Sun]{he2016deep}
Kaiming He, Xiangyu Zhang, Shaoqing Ren, and Jian Sun.
\newblock Deep residual learning for image recognition.
\newblock In \emph{Proceedings of the IEEE conference on computer vision and
  pattern recognition}, 2016.

\bibitem[He et~al.(2020)He, Fan, Wu, Xie, and Girshick]{he2020momentum}
Kaiming He, Haoqi Fan, Yuxin Wu, Saining Xie, and Ross Girshick.
\newblock Momentum contrast for unsupervised visual representation learning.
\newblock In \emph{Conference on Computer Vision and Pattern Recognition},
  2020.

\bibitem[Hitaj et~al.(2017)Hitaj, Ateniese, and P{\'e}rez-Cruz]{hitaj2017deep}
Briland Hitaj, Giuseppe Ateniese, and Fernando P{\'e}rez-Cruz.
\newblock Deep models under the gan: information leakage from collaborative
  deep learning.
\newblock In \emph{Proceedings of the 2017 ACM SIGSAC Conference on Computer
  and Communications Security}, 2017.

\bibitem[Ioffe \& Szegedy(2015)Ioffe and Szegedy]{ioffe2015batch}
Sergey Ioffe and Christian Szegedy.
\newblock Batch normalization: Accelerating deep network training by reducing
  internal covariate shift.
\newblock In \emph{International Conference on Machine Learning}, 2015.

\bibitem[Iyengar et~al.(2019)Iyengar, Near, Song, Thakkar, Thakurta, and
  Wang]{iyengar2019towards}
Roger Iyengar, Joseph~P Near, Dawn Song, Om~Thakkar, Abhradeep Thakurta, and
  Lun Wang.
\newblock Towards practical differentially private convex optimization.
\newblock In \emph{IEEE Symposium on Security and Privacy}, 2019.

\bibitem[Jayaraman et~al.(2018)Jayaraman, Wang, Evans, and
  Gu]{jayaraman2018distributed}
Bargav Jayaraman, Lingxiao Wang, David Evans, and Quanquan Gu.
\newblock Distributed learning without distress: Privacy-preserving empirical
  risk minimization.
\newblock In \emph{Advances in Neural Information Processing Systems}, 2018.

\bibitem[Jordon et~al.(2019)Jordon, Yoon, and van~der Schaar]{jordon2019pate}
James Jordon, Jinsung Yoon, and Mihaela van~der Schaar.
\newblock Pate-gan: Generating synthetic data with differential privacy
  guarantees.
\newblock In \emph{International Conference on Learning Representations}, 2019.

\bibitem[Kairouz et~al.(2020)Kairouz, Ribero, Rush, and
  Thakurta]{kairouz2020dimension}
Peter Kairouz, M{\'o}nica Ribero, Keith Rush, and Abhradeep Thakurta.
\newblock Dimension independence in unconstrained private erm via adaptive
  preconditioning.
\newblock \emph{arXiv preprint arXiv:2008.06570}, 2020.

\bibitem[Kifer et~al.(2012)Kifer, Smith, and Thakurta]{kifer2012private}
Daniel Kifer, Adam Smith, and Abhradeep Thakurta.
\newblock Private convex empirical risk minimization and high-dimensional
  regression.
\newblock In \emph{Conference on Learning Theory}, 2012.

\bibitem[Lee \& Kifer(2018)Lee and Kifer]{lee2018concentrated}
Jaewoo Lee and Daniel Kifer.
\newblock Concentrated differentially private gradient descent with adaptive
  per-iteration privacy budget.
\newblock In \emph{Proceedings of the 24th ACM SIGKDD International Conference
  on Knowledge Discovery \& Data Mining}, 2018.

\bibitem[Li et~al.(2020)Li, Gu, Zhou, Chen, and Banerjee]{li2020hessian}
Xinyan Li, Qilong Gu, Yingxue Zhou, Tiancong Chen, and Arindam Banerjee.
\newblock Hessian based analysis of sgd for deep nets: Dynamics and
  generalization.
\newblock In \emph{SIAM International Conference on Data Mining}, 2020.

\bibitem[Ma et~al.(2019)Ma, Zhu, and Hsu]{ma2019data}
Yuzhe Ma, Xiaojin Zhu, and Justin Hsu.
\newblock Data poisoning against differentially-private learners: attacks and
  defenses.
\newblock In \emph{Proceedings of the 28th International Joint Conference on
  Artificial Intelligence}, pp.\  4732--4738. AAAI Press, 2019.

\bibitem[Mironov(2017)]{mironov2017renyi}
Ilya Mironov.
\newblock R{\'e}nyi differential privacy.
\newblock In \emph{IEEE Computer Security Foundations Symposium}, 2017.

\bibitem[Mironov et~al.(2019)Mironov, Talwar, and Zhang]{mironov2019renyi}
Ilya Mironov, Kunal Talwar, and Li~Zhang.
\newblock R{\'e}nyi differential privacy of the sampled gaussian mechanism.
\newblock \emph{arXiv}, 2019.

\bibitem[Muirhead(2009)]{muirhead2009aspects}
Robb~J Muirhead.
\newblock \emph{Aspects of multivariate statistical theory}, volume 197.
\newblock John Wiley \& Sons, 2009.

\bibitem[Panju(2011)]{panju2011iterative}
Maysum Panju.
\newblock Iterative methods for computing eigenvalues and eigenvectors.
\newblock \emph{arXiv preprint arXiv:1105.1185}, 2011.

\bibitem[Papernot et~al.(2017)Papernot, Abadi, Erlingsson, Goodfellow, and
  Talwar]{papernot2016semi}
Nicolas Papernot, Mart{\'\i}n Abadi, Ulfar Erlingsson, Ian Goodfellow, and
  Kunal Talwar.
\newblock Semi-supervised knowledge transfer for deep learning from private
  training data.
\newblock In \emph{International Conference on Learning Representations}, 2017.

\bibitem[Papernot et~al.(2018)Papernot, Song, Mironov, Raghunathan, Talwar, and
  Erlingsson]{papernot2018scalable}
Nicolas Papernot, Shuang Song, Ilya Mironov, Ananth Raghunathan, Kunal Talwar,
  and {\'U}lfar Erlingsson.
\newblock Scalable private learning with pate.
\newblock In \emph{International Conference on Learning Representations}, 2018.

\bibitem[Phan et~al.(2020)Phan, Thai, Hu, Jin, Sun, and Dou]{phanscalable}
NhatHai Phan, My~T Thai, Han Hu, Ruoming Jin, Tong Sun, and Dejing Dou.
\newblock Scalable differential privacy with certified robustness in
  adversarial learning.
\newblock \emph{International Conference on Machine Learning}, 2020.

\bibitem[Rahman et~al.(2018)Rahman, Rahman, Laganiere, Mohammed, and
  Wang]{rahman2018membership}
Md~Atiqur Rahman, Tanzila Rahman, Robert Laganiere, Noman Mohammed, and Yang
  Wang.
\newblock Membership inference attack against differentially private deep
  learning model.
\newblock \emph{Transactions on Data Privacy}, 2018.

\bibitem[Sablayrolles et~al.(2019)Sablayrolles, Douze, Ollivier, Schmid, and
  J{\'e}gou]{sablayrolles2019white}
Alexandre Sablayrolles, Matthijs Douze, Yann Ollivier, Cordelia Schmid, and
  Herv{\'e} J{\'e}gou.
\newblock White-box vs black-box: Bayes optimal strategies for membership
  inference.
\newblock \emph{International Conference on Machine Learning}, 2019.

\bibitem[Shokri \& Shmatikov(2015)Shokri and Shmatikov]{shokri2015privacy}
Reza Shokri and Vitaly Shmatikov.
\newblock Privacy-preserving deep learning.
\newblock In \emph{ACM SIGSAC conference on computer and communications
  security}, 2015.

\bibitem[Shokri et~al.(2017)Shokri, Stronati, Song, and
  Shmatikov]{shokri2017membership}
Reza Shokri, Marco Stronati, Congzheng Song, and Vitaly Shmatikov.
\newblock Membership inference attacks against machine learning models.
\newblock In \emph{IEEE Symposium on Security and Privacy (SP)}, 2017.

\bibitem[Song et~al.(2013)Song, Chaudhuri, and Sarwate]{song2013stochastic}
Shuang Song, Kamalika Chaudhuri, and Anand~D Sarwate.
\newblock Stochastic gradient descent with differentially private updates.
\newblock In \emph{Global Conference on Signal and Information Processing
  (GlobalSIP)}, 2013.

\bibitem[Talwar et~al.(2015)Talwar, Thakurta, and Zhang]{talwar2015nearly}
Kunal Talwar, Abhradeep~Guha Thakurta, and Li~Zhang.
\newblock Nearly optimal private lasso.
\newblock In \emph{Advances in Neural Information Processing Systems}, 2015.

\bibitem[Thakurta \& Smith(2013)Thakurta and Smith]{thakurta2013differentially}
Abhradeep~Guha Thakurta and Adam Smith.
\newblock Differentially private feature selection via stability arguments, and
  the robustness of the lasso.
\newblock In \emph{Conference on Learning Theory}, 2013.

\bibitem[Tram{\`e}r \& Boneh(2020)Tram{\`e}r and
  Boneh]{tramer2020differentially}
Florian Tram{\`e}r and Dan Boneh.
\newblock Differentially private learning needs better features (or much more
  data).
\newblock \emph{arXiv preprint arXiv:2011.11660}, 2020.

\bibitem[Vogels et~al.(2019)Vogels, Karimireddy, and Jaggi]{vogels2019powersgd}
Thijs Vogels, Sai~Praneeth Karimireddy, and Martin Jaggi.
\newblock Powersgd: Practical low-rank gradient compression for distributed
  optimization.
\newblock In \emph{Advances in Neural Information Processing Systems}, 2019.

\bibitem[Wang \& Xu(2019)Wang and Xu]{wang2019sparse}
Di~Wang and Jinhui Xu.
\newblock On sparse linear regression in the local differential privacy model.
\newblock In \emph{International Conference on Machine Learning}, 2019.

\bibitem[Wang et~al.(2017)Wang, Ye, and Xu]{wang2017differentially}
Di~Wang, Minwei Ye, and Jinhui Xu.
\newblock Differentially private empirical risk minimization revisited: Faster
  and more general.
\newblock In \emph{Advances in Neural Information Processing Systems}, 2017.

\bibitem[Wang \& Zhou(2020)Wang and Zhou]{wang2020differentially}
Jun Wang and Zhi-Hua Zhou.
\newblock Differentially private learning with small public data.
\newblock In \emph{AAAI}, 2020.

\bibitem[Wang \& Gu(2019)Wang and Gu]{wang2019differentially}
Lingxiao Wang and Quanquan Gu.
\newblock Differentially private iterative gradient hard thresholding for
  sparse learning.
\newblock In \emph{International Joint Conference on Artificial Intelligence},
  2019.

\bibitem[Wang et~al.(2019)Wang, Balle, and Kasiviswanathan]{wang2019subsampled}
Yu-Xiang Wang, Borja Balle, and Shiva~Prasad Kasiviswanathan.
\newblock Subsampled r{\'e}nyi differential privacy and analytical moments
  accountant.
\newblock In \emph{International Conference on Artificial Intelligence and
  Statistics}, 2019.

\bibitem[Wu et~al.(2016)Wu, Fredrikson, Jha, and Naughton]{wu2016methodology}
Xi~Wu, Matthew Fredrikson, Somesh Jha, and Jeffrey~F Naughton.
\newblock A methodology for formalizing model-inversion attacks.
\newblock In \emph{IEEE Computer Security Foundations Symposium}, 2016.

\bibitem[Wu et~al.(2017)Wu, Li, Kumar, Chaudhuri, Jha, and
  Naughton]{wu2017bolt}
Xi~Wu, Fengan Li, Arun Kumar, Kamalika Chaudhuri, Somesh Jha, and Jeffrey
  Naughton.
\newblock Bolt-on differential privacy for scalable stochastic gradient
  descent-based analytics.
\newblock In \emph{ACM International Conference on Management of Data}, 2017.

\bibitem[Wu \& He(2018)Wu and He]{wu2018group}
Yuxin Wu and Kaiming He.
\newblock Group normalization.
\newblock In \emph{Proceedings of the European conference on computer vision
  (ECCV)}, 2018.

\bibitem[Yu et~al.(2020)Yu, Zhang, Chen, Yin, and Liu]{yu2020gradient}
Da~Yu, Huishuai Zhang, Wei Chen, Jian Yin, and Tie-Yan Liu.
\newblock Gradient perturbation is underrated for differentially private convex
  optimization.
\newblock In \emph{Proc. of 29th Int. Joint Conf. Artificial Intelligence},
  2020.

\bibitem[Yu et~al.(2021)Yu, Zhang, Chen, Yin, and Liu]{yu2021how}
Da~Yu, Huishuai Zhang, Wei Chen, Jian Yin, and Tie-Yan Liu.
\newblock How does data augmentation affect privacy in machine learning?
\newblock In \emph{Proc. of the AAAI Conference on Artificial Intelligence},
  2021.

\bibitem[Zhou et~al.(2020)Zhou, Wu, and Banerjee]{zhou2020bypassing}
Yingxue Zhou, Zhiwei~Steven Wu, and Arindam Banerjee.
\newblock Bypassing the ambient dimension: Private sgd with gradient subspace
  identification.
\newblock \emph{arXiv preprint arXiv:2007.03813}, 2020.

\bibitem[Zhu et~al.(2019)Zhu, Liu, and Han]{zhu2019deep}
Ligeng Zhu, Zhijian Liu, and Song Han.
\newblock Deep leakage from gradients.
\newblock In \emph{Advances in Neural Information Processing Systems}, 2019.

\bibitem[Zhu \& Wang(2019)Zhu and Wang]{zhu2019poission}
Yuqing Zhu and Yu-Xiang Wang.
\newblock Poission subsampled r{\'e}nyi differential privacy.
\newblock In \emph{International Conference on Machine Learning}, 2019.

\end{thebibliography}
\bibliographystyle{iclr2021_conference}

\newpage

\appendix

\section{Experiments with pre-trained models}
\label{sec:linear}

Recent works have shown that pre-training the models on unlabeled data can be beneficial for subsequent learning tasks \citep{chen2020big,he2020momentum}. \cite{tramer2020differentially} demonstrate that differentially private linear classifier can achieve high accuracy using the features produced by those per-trained models.  We show that GEP can also benefit from such pre-trained models.

Inspired by \cite{tramer2020differentially}, we use the output of the penultimate
layer of a pre-trained ResNet152 model  as feature to train a private linear classifier. The ResNet152 model is pre-trained on unlabeled ImageNet  using SimCLR \citep{chen2020big}.  The feature dimension is 4096.

\textbf{Implementation Details} We choose the privacy parameter $\epsilon$ from $[0.1, 0.5, 1, 2]$. The privacy parameter $\delta$ is $1\times 10^{-5}$.  We run all experiments for 5 times and report the average accuracy.  The clipping threshold of residual gradients is still one-fifth of the clipping threshold of the original gradients. The dimension of anchor subspace is set as $200 \simeq\sqrt{p}$ where $p=40960$ is the model dimension. We randomly sample $500$ samples from the test set as auxiliary data and evaluate performance on the rest test samples.   The optimizer is Adam with default momentum coefficients.  Other hyper-parameters are listed in Table~\ref{tbl:hyper}.

\begin{table}  [h]
\centering
\begin{tabular}{|c|c|}

\hline \hline
Hyperparameter					& Values	\\	\hline	
Learning rate			 & 0.01, 0.05, 0.1			\\	\hline
Running steps				& 50, 100, 400 \\ \hline
Clipping  threshold			& 0.01, 0.1, 1 \\ \hline

\hline
\hline
\end{tabular}
\caption{Hyperparameter values used in Appendix~\ref{sec:linear}. }
\label{tbl:hyper}
\end{table}

\textbf{Results} The experiment results are shown in Table~\ref{tbl:five_eps0.5}. GEP outperforms GP on all values of $\epsilon$. With privacy bound $\epsilon=2$, GEP achieves 94.8\% validation accuracy on CIFAR10 dataset, improving over the GP baseline by 1.4\%. For very strong privacy guarantee ($\epsilon=0.1$), B-GEP performs on par with GEP because strong privacy guarantee requires large noise and the useful signal in residual gradient is submerged in the added noise. B-GEP benefits less from larger $\epsilon$ compared to GP or GEP. For $\epsilon=1$ and $2$, the performance of B-GEP is worse than the performance of GP. This is because larger $\epsilon$ can not reduce the systematic error of B-GEP (see Remark~\ref{thm:error_cp} in Section~\ref{sec:ppgr_error}).

\begin{table} [h]
    \caption{Validation accuracy (in \%) on CIFAR10 with varying choices of $\epsilon$. We train a private linear model on top of the features from a ResNet152 model, which is pre-trained on unlabeled ImageNet. }
\label{tbl:five_eps0.5}
\centering
\begin{tabular}{ p{1.75cm}p{2.25cm}p{2.25cm}p{2.25cm} p{2.25cm}}
 \hline \hline
               & $\epsilon=0.1$ 	& $\epsilon=0.5$  		& $\epsilon=1$ 		& $\epsilon=2$ \\[0.4ex]
 \hline
Non private    &   	96.3   & 		96.3     &  	96.3		  &   		96.3	\\[0.4ex]
 GP		      &  88.2 ($\pm$0.16)	 	 &  91.1 ($\pm$0.17)  	 & 	 93.2 ($\pm$0.19)		 & 	 93.4 ($\pm$0.12)		 \\[0.4ex]
B-GEP   &   \textbf{91.0} ($\pm$0.07)	 &     92.9 ($\pm$0.03) & 	93.1 ($\pm$0.10)		&  	93.2 ($\pm$0.08)	\\[0.4ex] 
GEP	   &  		90.9 ($\pm$0.19)	&    \textbf{93.5} 	($\pm$0.06) &    	\textbf{94.3} ($\pm$0.09)	& 	 \textbf{94.8} ($\pm$0.06)	\\[0.4ex]	

 \hline
 \hline
\end{tabular} 
\end{table}

\section{Complexity Analysis}
\label{sec:complexity}

We provide an analysis of the computational and memory costs of the construction of anchor subspace. The computation of the anchor subspace is the dominant additional cost of GEP compared to conventional gradient perturbation. Notations: $k$, $m$, $n$, and $p$ are the dimension of anchor subspace, number of anchor gradients, number of private gradients, and the model dimension, respectively.  In order to reduce the computational and memory costs, we divide the parameters into $g$ groups and compute one orthonormal basis for each group. We refer to this approach as `parameter grouping'. In this section, we assume the parameters and the dimension of the anchor subspace are both divided evenly. Table~\ref{tbl:cost} summarizes the additional costs of GEP with/without parameter grouping. Using parameter grouping can  reduce the computational/memory cost significantly.

\begin{table} 
    \caption{Computational and memory costs of a single power iteration in Algorithm~\ref{alg:ppg}. The computation cost is measured by the number of floating point operations. The memory cost is measured by the number of floating-point numbers we need to store. `GEP+PG' denotes GEP with parameter grouping and $g$ denotes the number of groups. Notations: $k$, $m$, $n$, and $p$ are the dimension of anchor subspace, number of anchor gradients, number of private gradients, and the model dimension, respectively. }
\label{tbl:cost}
\centering

\begin{tabular}{ p{1.75cm}p{5cm}p{5cm} }
 \hline \hline
               & Computational Cost 	& Memory Cost 		 \\[0.4ex]
 \hline
GEP    &   	$2mkp+pk^2$   & 		$\max\left(0,\left(m-n+k\right)p + mk\right)$    			\\[0.4ex]
GEP+PG	      &  $2mkp/g+pk^2/g^2$	 	 &  $\max\left(0,\left(m-n+\frac{k}{g}\right)p + mk\right)$ 		 \\[0.4ex]

 \hline
 \hline
\end{tabular} 
\end{table}

\section{Ablation Study}
\label{apdx:abla}

\textbf{The influence of choosing different auxiliary datasets.} We conduct experiments with different choices of auxiliary datasets. For CIFAR10, we try 2000 random test samples from CIFAR10, 2000 random samples from CIFAR100, and 2000 random samples from ImageNet. When the auxiliary dataset is CIFAR10, we try both correct labels and random labels. For all choices of auxiliary datasets, the test accuracy is evaluated on 8000 test samples of CIFAR10 that are not used as auxiliary data.  Other implementation details are the same as in Section~\ref{sec:exp}. The results are shown in Table~\ref{tbl:aux_data}. Surprisingly, using samples from CIFAR10 with correct labels yields the worst accuracy. This may because  the model  `overfits'  the auxiliary data when it has access to correct labels, which makes the anchor subspace contains less information about the private gradients. The best accuracy is achieved using samples from CIFAR10 with random labels, this makes sense because in this case the features of auxiliary data and private data have the same distribution. Using samples from CIFAR100 or ImageNet as auxiliary data  has a small influence on the test accuracy.

\begin{table} [h]
    \caption{Test accuracy on CIFAR10 with different choices of auxiliary datasets. The privacy guarantee is $(8,10^{-5})$-DP. We report the average accuracy of five runs with standard deviations  in brackets.  }
\label{tbl:aux_data}
\centering
\begin{tabular}{ p{3.5cm}p{3.5cm}p{3.5cm}}
 \hline \hline
Auxiliary Data   & Random Label? 	& Test Accuracy  	 \\[0.4ex]
 \hline
CIFAR10    &   	No  & 		72.9 ($\pm$0.31)    	\\[0.4ex]
 CIFAR10	 &  Yes	 &  \textbf{75.1} ($\pm$0.42)  \\[0.4ex]
CIFAR100   &  Yes &  74.7 ($\pm$0.46)\\[0.4ex] 
ImageNet  &  Yes & 74.8  ($\pm$0.39)	\\[0.4ex]	

 \hline
 \hline
\end{tabular} 
\end{table}

\textbf{The influence of the number of anchor gradients.} In the main text, the size of auxiliary dataset is $m=2000$. We conduct more experiments with different sizes of auxiliary dataset to examine the influence of $m$. The auxiliary data is randomly sampled from ImageNet. Table~\ref{tbl:aux_data_size} reports the test accuracy on CIFAR10 with different choices of $m$.  For both B-GEP and GEP, increasing  $m$  leads to slightly improved performance.

\begin{table} [h]
    \caption{Test accuracy on CIFAR10 with different sizes of auxiliary dataset. The privacy guarantee is $(8,10^{-5})$-DP. We report the average accuracy of five runs with standard deviations  in brackets.  }
\label{tbl:aux_data_size}
\centering
\begin{tabular}{ p{2.5cm}p{2.5cm}p{2.5cm}p{2.5cm}}
 \hline \hline
Algorithm   & $m=1000$ 	& $m=2000$ 	& $m=4000$ \\[0.4ex]
 \hline
B-GEP    &   62.2 ($\pm$0.26)	  & 62.6 ($\pm$0.24)  & 63.3 ($\pm$0.27)	   	\\[0.4ex]
GEP	 &  74.6 ($\pm$0.41)	 &  74.8 ($\pm$0.39) &  \textbf{75.2} ($\pm$0.34) 	  \\[0.4ex]

 \hline
 \hline
\end{tabular} 
\end{table}

\begin{figure}
    \centering
  \includegraphics[width=0.6\linewidth]{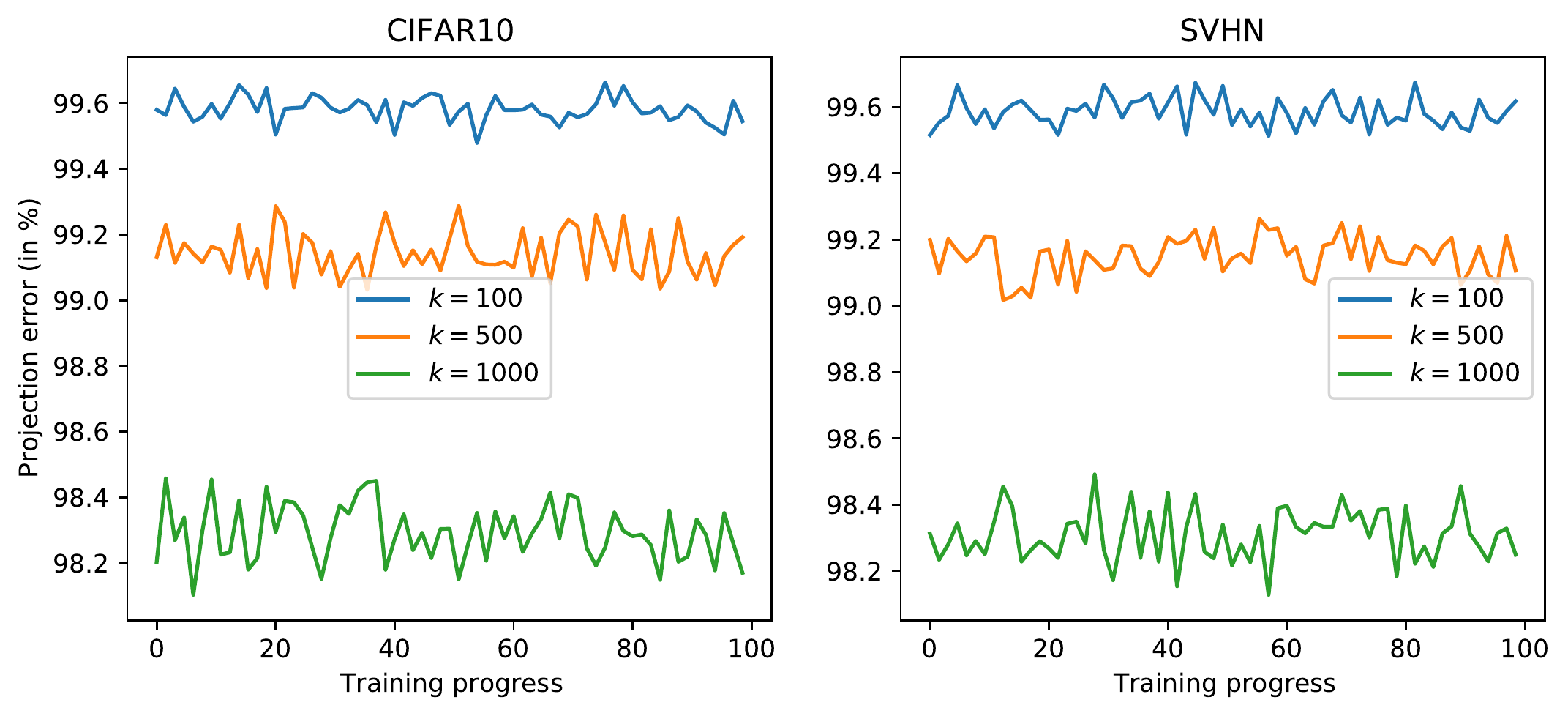}
  \caption{Projection error rate of random basis vectors.  The dimension of  subspace is denoted by $k$.  }
  \label{fig:rp_projection_error}
\end{figure}

\textbf{The projection error of random basis vectors.} It is tempting to construct the anchor subspace using random basis vectors because Johnson–Lindenstrauss Lemma \citep{dasgupta2003elementary}  guarantees that one can preserve the pairwise distance between any two points after projecting into a random subspace of much lower dimension. We empirically verify the projection error of Gaussian random basis vectors on CIFAR10 and SVHN. The experiment settings are the same as in Section~\ref{sec:exp}. The projection errors over the training process are plotted in Figure~\ref{fig:rp_projection_error}. The projection error of random basis vectors is very high ($>95\%$) throughout training. This is because  preserving the pairwise distance is not sufficient for high quality gradient reconstruction, which requires one to preserve the average ‘distance’ between any individual gradient and all other gradients.

\section{Missing Proofs}
\label{sec:proof_dp}

\lowrankerror*

\begin{proof}
We extend the Theorem 3.2 in \cite{eaton1973non} to the low-rank case.
\begin{theorem}[Theorem 3.2 in \cite{eaton1973non}] \label{thm:non-singularity}
Let $\mX=(\vx_1, ..., \vx_n)$ where the $\vx_i$ are i.i.d. random vectors in $\sR^p$, $n\ge p$. If $\sP\{\vx_1 \in \sM\}=0$ for all proper manifolds $\sM\subset \sR^p$, then $\sP\{\mX \text{ is non-singular}\}$=1.
\end{theorem}
We note that the subspace spanned by $\hat{\mV}_{k'}$ is in the space spanned by $\mV_{k}$ by definition. Hence $k'\le k$.

Let $\hat{\vx}_i := \mV_k^T \hat{\vxi}_i \in \sR^k$ for $i\in [m]$. Then $\hat{\mX} := (\hat\vx_1, ..., \hat\vx_m)$ is non-singular because of the assumption and Theorem \ref{thm:non-singularity}. That is $rank(\hat{\mX}) =k$. Therefore $rank((\hat\vxi_1, ..., \hat\vxi_m))\ge k$, $rank(\hat\mS)\ge k$ and $k'\ge k$. Therefore $k'=k$ and the subspace spanned by $\hat{\mV}_{k'}$ and the subspace spanned by $\mV_{k}$ are identical.
\end{proof}

\privacyppgr*

\begin{proof}[Proof of Theorem~\ref{thm:privacy_ppgr}]

 We first introduce some background knowledge of Rényi differential privacy (RDP)  \citep{mironov2017renyi}. RDP measures the Rényi divergence between two output distributions.

\begin{definition}[$(\lambda,\gamma)$-RDP]
A randomized mechanism $f$ is said to guarantee $(\lambda,\gamma)$-RDP if for any neighboring datasets $\sD,\sD'$ and $\lambda>1$ it holds that  \[D_{\lambda}(f(\sD)||f(\sD'))\leq\gamma,\]
where $D_{\lambda}(\cdot||\cdot)$ denotes the Rényi divergence of order $\lambda$.
\end{definition}

We next introduce some useful properties of RDP.

\begin{lemma} [Gaussian mechanism of RDP]
\label{lma:gm_rdp}
Let $S=\max_{\sD\sim \sD'}\norm{f(\sD)-f(\sD')}$ be the $l_{2}$ sensitivity, then Gaussian mechanism $\mathcal{M}=f(\sD)+\vz$ satisfies $(\lambda,\frac{\lambda S^{2}}{2\sigma^{2}})$-RDP, where $\vz\sim\mathcal{N}(0,\sigma^{2}I_{p\times p})$.
\end{lemma}

\begin{lemma} [Composition of RDP]
\label{lma:composition}
If $M_{1}$, $M_{2}$ satisfy $(\lambda,\gamma_{1})$-RDP and $(\lambda,\gamma_{2})$-RDP respectively, then their composition satisfies $(\lambda,\gamma_{1}+\gamma_{2})$-RDP.
\end{lemma}

\begin{lemma} [Conversion from RDP to $(\epsilon,\delta)$-DP]
\label{lma:conversion}
If $\mathcal{M}$ obeys $(\lambda,\gamma)$-RDP, then $\mathcal{M}$ obeys $(\gamma+\log(1/\delta)/(\lambda-1), \delta)$-DP for all $0<\delta<1$.
\end{lemma}

%and some properties of RDP. 
\medskip
Now we proof Theorem~\ref{thm:privacy_ppgr}. Let $\mW,\mW'$ be the gradient embeddings of two neighboring datasets $\sD\sim \sD'$ and $\mR,\mR'$ be corresponding residual gradients. Without loss of generality, suppose $\mW$ ($\mR$) has one more row than $\mW'$ ($\mR'$). For given sensitivity $S_{1}, S_{2}$, 
\[\max_{\sD\sim \sD'}\norm{\vw-\vw'}=\max_{\mW\sim \mW'}\norm{ \mW_{n,:}}\leq S_{1}, \quad \max_{\sD\sim \sD'}\norm{\vr-\vr'}=\max_{ \mR\sim \mR'}\norm{ \mR_{n,:}}\leq S_{2}.\]

If we set $\sigma_{1}=S_{1}\sigma$ and $\sigma_{2}=S_{2}\sigma$ for some $\sigma$, then Algorithm~\ref{alg:ppg} satisfies $(\lambda,\frac{\lambda}{\sigma^{2}})$-RDP because of Lemma~\ref{lma:gm_rdp} and~\ref{lma:composition}. In order to guarantee $(\epsilon,\delta)$-DP, we need
\begin{equation}
\begin{aligned}
\label{eq:privacyppg0}
\frac{\lambda }{\sigma^{2}} + \frac{\log(1/\delta)}{\lambda-1} \leq \epsilon.
\end{aligned}
\end{equation}
Choose $\lambda=1+\frac{2\log(1/\delta)}{\epsilon}$ and rearrange Eq~(\ref{eq:privacyppg0}),  we need
\begin{equation}
\begin{aligned}
\label{eq:privacyppg1}
\sigma^{2}\geq \frac{2\left(\epsilon+2\log(1/\delta)\right)}{\epsilon^2}.
\end{aligned}
\end{equation}

Then using the constraint on $\epsilon$ concludes the proof.

\end{proof}

\privacygd*

\begin{proof}[Proof of Theorem~\ref{thm:privacy_gd}]
From the proof of Theorem~\ref{thm:privacy_ppgr}, we have each call of GEP satisfies $(\lambda,\frac{\lambda}{\sigma^{2}})$-RDP. Then by the composition property of RDP (Lemma~\ref{lma:composition}), the output of Algorithm~\ref{alg:dp_gd} satisfies $(\lambda,\frac{T\lambda}{\sigma^{2}})$-RDP. Plugging $\frac{T\lambda}{\sigma^{2}}$ into Equation~\ref{eq:privacyppg0} and~\ref{eq:privacyppg1} concludes the proof.

\end{proof}

\convergence*

\begin{proof} [Proof of Theorem~\ref{thm:convergence}]
The $\beta$-smooth condition gives
\begin{equation}
\begin{aligned}
\label{eq:convergence_eq0}
L(\boldsymbol\theta_{t+1})\leq L(\boldsymbol\theta_{t})+\idot{\nabla L(\boldsymbol\theta_{t}),\boldsymbol\theta_{t+1}-\boldsymbol\theta_{t}}+\frac{\beta}{2}\norm{\boldsymbol\theta_{t+1}-\boldsymbol\theta_{t}}^{2}.
\end{aligned}
\end{equation}

\newcommand{\bt}{\boldsymbol\theta}

Based on the update rule of GEP  we have
\begin{equation}
\begin{aligned}
\label{eq:convergence_eq1}
\boldsymbol\theta_{t+1}-\boldsymbol\theta_{t}=-\eta\tilde\vv=-\eta\nabla L(\bt_{t})-\frac{\eta}{n}(\vz^{(1)}_{t}\mB+\vz^{(2)}_{t}),
\end{aligned}
\end{equation}
where $\vz^{(1)}_{t}\sim\mathcal{N}(0,\sigma^{2}\mI_{k\times k})$,   $\vz^{(2)}_{t}\sim\mathcal{N}(0,\sigma^{2}r_{t}^{2}\mI_{p\times p})$ are the perturbation noises  and $r_{t}=\max_{i} \norm{(\mR_{t})_{i,:}}$ is the sensitivity of residual gradients at step $t$.

Take expectation on Eq~(\ref{eq:convergence_eq0}) with respect to the perturbation noises.
\begin{equation}
\begin{aligned}
\label{eq:convergence_eq2}
\mathbb{E}[L(\boldsymbol\theta_{t+1})]\leq \mathbb{E}[L(\boldsymbol\theta_{t})]-(\eta-\beta\eta^{2}/2)\mathbb{E}[\norm{\nabla L(\boldsymbol\theta_{t})}^{2}]+\frac{\beta\eta^{2}\sigma^{2}}{2 n^2}\left(k+pr_{t}^{2}\right).
\end{aligned}
\end{equation}
Subtract $L(\bt_{*})$ from both sides, we have
\begin{equation}
\begin{aligned}
\label{eq:convergence_eq3}
\mathbb{E}[L(\boldsymbol\theta_{t+1})] - L(\bt_{*})&\leq \mathbb{E}[L(\boldsymbol\theta_{t})] - L(\bt_{*})-(\eta-\beta\eta^{2}/2)\mathbb{E}[\norm{\nabla L(\boldsymbol\theta_{t})}^{2}]+\frac{\beta\eta^{2}\sigma^{2}}{2 n^2}\left(k+pr_{t}^{2}\right)\\
&\leq \mathbb{E}[\idot{\nabla L(\bt_{t}),\bt_{t}-\bt_{*}}]-(\eta-\beta\eta^{2}/2)\mathbb{E}[\norm{\nabla L(\boldsymbol\theta_{t})}^{2}]+\frac{\beta\eta^{2}\sigma^{2}}{2 n^2}\left(k+pr_{t}^{2}\right).
\end{aligned}
\end{equation}
The second inequality holds because $L$ is convex. Then choose $\eta=\frac{1}{\beta}$ and plug $\nabla L(\boldsymbol\theta_{t})=(\bt_{t}-\bt_{t+1})/\eta- (\vz^{t}_{1}\mB+\vz^{t}_{2})/n$ into Eq~(\ref{eq:convergence_eq3}).

\begin{equation}
\begin{aligned}
\label{eq:convergence_eq4}
\mathbb{E}[L(\boldsymbol\theta_{t+1})] - L(\bt_{*})&\leq \beta\mathbb{E}[\idot{\bt_{t}-\bt_{t+1},\bt_{t}-\bt_{*}}]-\frac{\beta}{2}\mathbb{E}[\norm{\bt_{t}-\bt_{t+1}}^{2}]+\frac{\sigma^{2}}{\beta n^2}\left(k+pr_{t}^{2}\right)\\
&=\frac{\beta}{2}\left(\mathbb{E}[\norm{\bt_{t}-\bt_{*}}^2]-\mathbb{E}[\norm{\bt_{t+1}-\bt_{*}}^{2}]\right)+\frac{\sigma^{2}}{\beta n^2}\left(k+pr_{t}^{2}\right).
\end{aligned}
\end{equation}

Sum over $t=0,\ldots,T-1$ and use convexity, we have
\begin{equation}
\begin{aligned}
\label{eq:convergence_eq5}
\mathbb{E}[L(\boldsymbol{\bar\theta})]-L(\boldsymbol{\theta_{*}})\leq 
\frac{\beta}{2T}\norm{\bt_{0}-\bt_{*}} + \frac{\sigma^{2}}{\beta n^{2}}(k+\frac{p}{T}\sum_{t=0}^{T-1}r_{t}^{2}).
\end{aligned}
\end{equation}

Then substituting $T=\frac{n\beta\epsilon}{\sqrt{p}}$ and $\sigma=\mathcal{O}(\sqrt{T\log(1/\delta)}/\epsilon)$ yields the desired bound.

\end{proof}
%\appendix
% \section{Appendix}

\end{document}